\documentclass[journal]{IEEEtran}

\ifCLASSINFOpdf
\else
\fi
\usepackage{url}
\usepackage{color,soul}
\usepackage{epsfig}
\usepackage{graphicx}
\usepackage{amsmath,amsfonts,amsthm,amssymb}
\usepackage{arydshln}
\usepackage{subcaption,cite}
\usepackage{algorithm}
\usepackage{algpseudocode}
\usepackage{algorithmicx}
\usepackage{float}
\usepackage{adjustbox,balance}
\usepackage{multirow,capt-of, etoolbox}
\newcommand{\ie}{i.e.~\!}
\newcommand{\etal}{et al.~\!}
\newcommand{\eg}{e.g.~\!}



\begin{document}

\title{Improved ArtGAN for Conditional Synthesis of Natural Image and Artwork}

\author{Wei Ren Tan,~\IEEEmembership{Student Member,~IEEE,}
        Chee Seng Chan,~\IEEEmembership{Senior~Member,~IEEE,}\\
        Hern\'an E. Aguirre,~\IEEEmembership{Member,~IEEE,}
        and Kiyoshi Tanaka, \IEEEmembership{Member, IEEE}
\thanks{Manuscript received August 30, 2017; revised July 14, 2018; accepted on August 13, 2018. This work is supported in part by the Fundamental Research Grant Scheme (FRGS) MoHE Grant FP004-2016, from the Ministry of Education Malaysia and in part by the UM Frontier Research Grant FG002-17AFR, from University of Malaya. The associate editor coordinating the review of this manuscript and approving it for publication was Dr. Catarina Brites. \textit{(Corresponding author: Chee Seng Chan)}}
\thanks{W.R. Tan, A. Hernan and K. Tanaka are with Shinshu University, Nagano, Japan. e-mail: \{14st203c,ahernan,ktanaka\}@shinshu-u.ac.jp}
\thanks{C.S. Chan is with Center of Image and Signal Processing, Faculty of Computer Science and Information Technology, University of Malaya, Kuala Lumpur, Malaysia. e-mail: cschan@um.edu.my}
}

\markboth{}%
{Tan \MakeLowercase{\textit{\etal}}: ArtGAN}



\makeatletter
\let\@oldmaketitle\@maketitle
\renewcommand{\@maketitle}{\@oldmaketitle
\centering
  \includegraphics[width=\linewidth]{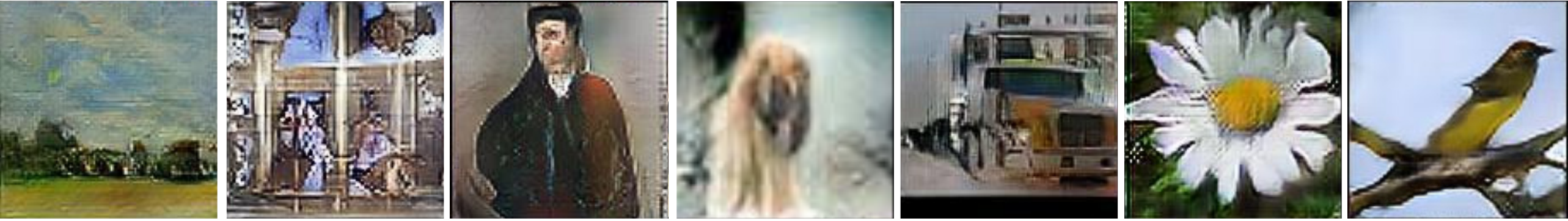}}
\makeatother

\maketitle 

\begin{abstract}
This paper proposes a series of new approaches to improve Generative Adversarial Network (GAN) for conditional image synthesis and we name the proposed model as ``ArtGAN". One of the key innovation of ArtGAN is that, the gradient of the loss function w.r.t. the label (randomly assigned to each generated image) is back-propagated from the categorical discriminator to the generator. With the feedback from the label information, the generator is able to learn more efficiently and generate image with better quality. Inspired by recent works, an autoencoder is incorporated into the categorical discriminator for additional complementary information. Last but not least, we introduce a novel strategy to improve the image quality. In the experiments, we evaluate ArtGAN on CIFAR-10 and STL-10 via ablation studies. The empirical results showed that our proposed model outperforms the state-of-the-art results on CIFAR-10 in terms of Inception score. Qualitatively, we demonstrate that ArtGAN is able to generate plausible-looking images on Oxford-102 and CUB-200, as well as able to draw realistic artworks based on style, artist, and genre. The source code and models are available at: \url{https://github.com/cs-chan/ArtGAN}.

\end{abstract}


\IEEEpeerreviewmaketitle
\section{Introduction}

Recently, Goodfellow \etal \cite{goodfellow2014generative} proposed an interesting features learning model called Generative Adversarial Networks (GAN) by employing two neural networks that are adversarially trained. Unlike the traditional deep discriminative models \cite{simonyan2014very, szegedy2015going, he2016deep}, the representations learned by GAN can be visualized through the generator in GAN in the form of synthetic images. More interestingly, these generated images look more realistic to human observers compared to other generative models. Since then, many extensions of GAN \cite{mirza2014conditional, radford2015unsupervised, springenberg2015unsupervised, denton2015deep, odena2016conditional, reed2016generative, zhang2016stackgan} have been introduced and showed promising results in generating appealing images when trained on datasets, such as MNIST \cite{lecun1998mnist}, CIFAR-10 \cite{krizhevsky2009learning}, ImageNet \cite{ILSVRC15}, etc. Despite the success, there is still room for improvement as the synthetic image quality is still far from realistic.

While unconditional GAN is an important research area, this paper is interested in class-conditioned GAN. In particular, conditional GAN is useful to understand how the visual representation of each class is learned via the visualization techniques inherent in GAN. Furthermore, we are interested to investigate if a machine can generate artwork based on style, genre, or artist. Artwork is a mode of creative expression, coming in different kinds of forms, including drawing, naturalistic, abstraction, etc. Unlike the aforementioned datasets \cite{lecun1998mnist, krizhevsky2009learning, ILSVRC15}, the representations of artworks can be harder to learn because they are usually non-figurative or abstract. 

\begin{figure*}[t]
\centering
\includegraphics[width=0.9\linewidth, height=0.3\linewidth]{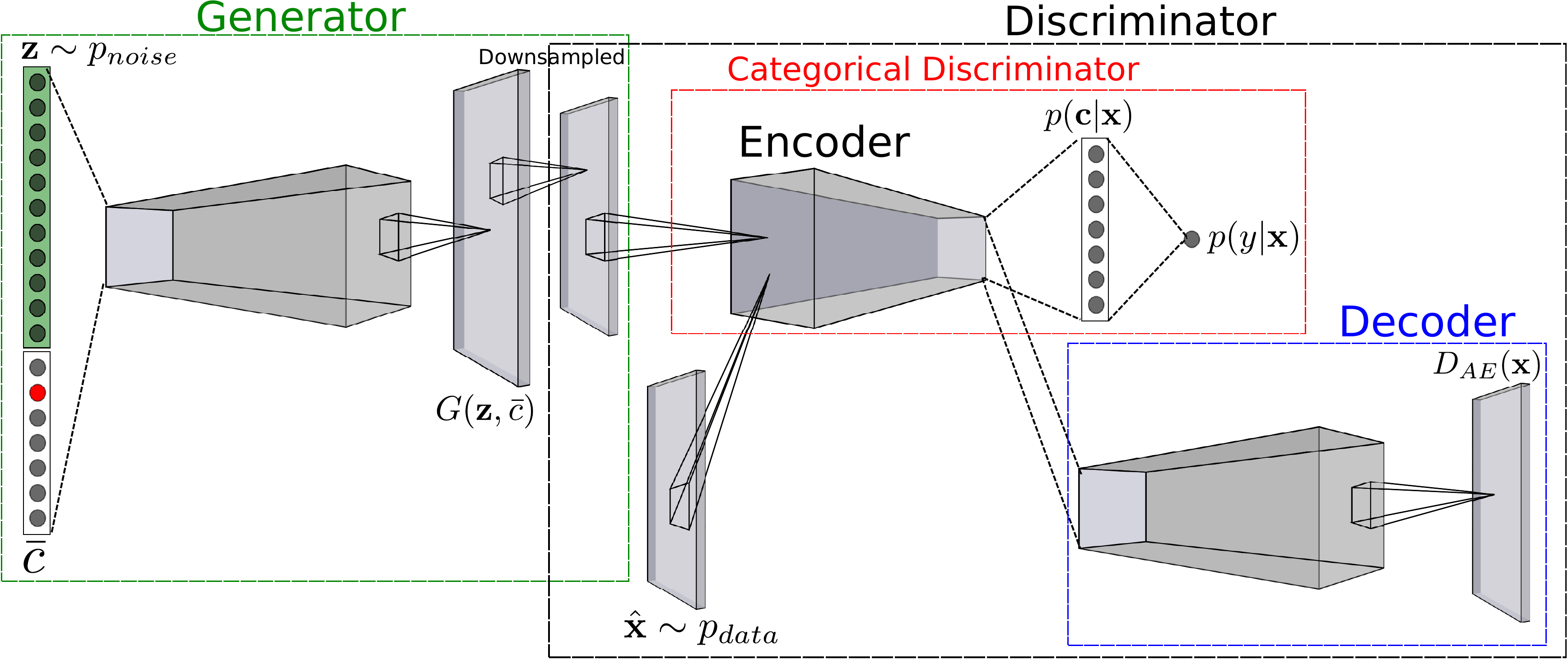}
\caption{Overview of ArtGAN-AEM architecture. $\mathbf{z}$ and $\hat{c}$ are concatenated and fed to the generator to produce synthetic image $G(\mathbf{z}, \hat{c})$. Either the downsampled generated image $G(\mathbf{z}, \hat{c})$ or real data $\hat{\mathbf{x}}$ is used as the input $\mathbf{x}$ to the (categorical autoencoder-based) discriminator. The discriminator produces three outputs: the class prediction $p(\mathbf{c}|\mathbf{x})$, adversarial prediction $p(y|\mathbf{x})$, and the reconstructed image $D_{AE}(\mathbf{x})$.}\vspace{-.1in}
\label{archview}
\end{figure*}

To this end, we propose a novel conditional GAN named \textbf{ArtGAN} for conditional synthesis of natural image and artwork. We anticipate that a good way to look at this problem is to understand how humans learn to draw. An artist teacher wrote an online article\footnote{http://www.learning-to-see.co.uk/effective-practice} and pointed out that an effective learning requires to focus on a particular type of skills at a time, \eg~practice to draw a particular object or one kind of movement at a time. Accordingly, ArtGAN takes a randomly chosen label information and a noise vector as inputs. The chosen label is used as the true label when computing the loss function for the generated image. The idea is to allow the generator to learn more efficiently by leveraging the feedback information from the labels. Inspired by recent works \cite{zhao2016energy, berthelot2017began}, a categorical autoencoder-based discriminator that incorporates an autoencoder into the categorical discriminator for additional complementary information is introduced. Rather than deploying two separate computationally expensive networks (\ie~a categorical discriminator and an autoencoder separately), the categorical autoencoder-based discriminator in our proposed GAN partly shares the same architecture and weights. In specific, \textit{encoder} in the autoencoder is shared by the categorical discriminator as illustrated in Figure \ref{archview}. 

In addition, we introduce a novel strategy to improve the generated image quality. The motivation behind this strategy is to generate a set of pixels that vote for a better quality pixel via average ranking in order to generate better pixel values. One may naively train an ensemble GANs to achieve this goal. However, training multiple GANs explicitly is computationally expensive and does not guarantee to achieve similar performance gain \cite{wang2016ensembles, huang2017snapshot}. Hence, we innovate an alternative approach where the generator in ArtGAN will generate synthetic images with resolution $2\times$ higher than the original image size. Then, these generated images will be downsampled to the original size using averaged pooling operation as a form of voting scheme.

In summary, our key contributions are: i) We propose ArtGAN to emulate the concept of effective learning to generate very challenging images. Within this, we introduce a novel way to improve the image quality. ii) Empirically, we show that our proposed models are able to generate CIFAR-10 \cite{krizhevsky2009learning} and STL-10 \cite{coates2011analysis} images with better Inception scores compared to the state-of-the-art results. iii) Our models are capable of generating Oxford-102 \cite{Nilsback08} and CUB-200 \cite{WahCUB_200_2011} samples that contain clear object structures in them. At the same time, ArtGAN is also able to generate high quality artwork that exhibit similar visual representations within \textit{genre}, \textit{artist}, or \textit{style}. To the best of our knowledge, no existing empirical research has addressed the implementation of a generative model on a large scale artworks dataset. 

A preliminary version of this work was presented earlier \cite{tan2017artgan}. The present work adds to the initial version in significant ways. First, we extend ArtGAN with the introduction of categorical autoencoder-based discriminator. Secondly, we innovate a way to improve the image quality generated by ArtGAN. Thirdly, considerable new analysis and intuitive explanations are added to the initial results. For instance, we extend the original qualitative experiments from Wikiart \cite{saleh2015large} to CIFAR-10 \cite{krizhevsky2009learning}, STL-10 \cite{coates2011analysis}, Oxford-102 \cite{Nilsback08}, and CUB-200 \cite{WahCUB_200_2011} datasets. In addition, we included the Inception score \cite{salimans2016improved} as a quantitative metric where ArtGAN obtained state-of-the-art result on CIFAR-10 dataset. 

The rest of the paper is structured as follows. Related works are discussed in the next section (Section \ref{secrelated}). Section \ref{secP} describes the proposed models, while the image quality strategy is explained in detail in Section \ref{secml}. Experiments are discussed in Section \ref{secex}. Last but not least, conclusion is drawn in Section \ref{secconclude}. 

\section{Related works}
\label{secrelated}

Generative models have been a fundamental interest and challenging problem in the field of computer vision and machine learning. In contrast to discriminative models which only allow sampling of the target variables conditioned on the observed quantities, generative models can be used to simulate observed distribution, and so they offer a much richer representation. Early works \cite{mnih2010generating, le2011learning, shim2012probabilistic} studied the statistical properties of natural images, but are limited to texture or certain patterns (\eg faces) only due to the difficulty in learning an effective feature representation. Recently, advances in deep models nourish a series of deep generative models \cite{kingma2013auto, kingma2014semi} for image generation through the Bayesian inference, typically trained by maximizing the log-likelihood. These models are able to construct decent quality images on less complicated images, such as digits and faces, but generally have intractable likelihood and require numerous approximations. Denoising autoencoders (DAE) \cite{bengio2013generalized} were introduced to overcome the intractable problem, but the reconstructed images are generally blurry. Then, DRAW \cite{gregor2015draw} was proposed, depicted as a sequential model with attention mechanism to draw image recursively. It mimics the process of human drawing but faces challenges when it is scaled up to large and complex images. PixelRNN \cite{oord2016pixel} is another autoregressive approach for image generation that has received much attentions recently. Its extensions (PixelCNN \cite{van2016conditional} and PixelCNN++ \cite{salimans2017pixelcnn++}) are able to synthesize decent images but are computationally expensive to train\footnote{They reported that PixelCNN++ requires approximately 5 days to converge to the reported results using 8 Maxwell TITAN X GPUs in github: \url{https://github.com/openai/pixel-cnn}.}.

Recently, a more significant breakthrough framework, Generative Adversarial Network (GAN) was introduced by Goodfellow \etal~\cite{goodfellow2014generative}. This framework escapes the difficulty of maximum likelihood estimation by estimating the generative model via an adversarial process and has gained striking successes in natural image generation. However, GAN is well-known for its instability during training. To tackle this problem,  feature matching \cite{salimans2016improved} was proposed to generate descent quality images. Instance noise \cite{sonderby2016amortised} is also an effective method to remedy the instability problem. Several variants proposed to address this problem by analysing the objective function of GAN. Wasserstein GAN (WGAN) used the Lipschitz constrained Earth-Mover (EM) distance to address the vanishing gradient and the saturated Jensen-Shannon distance problems. However, WGAN can still generate low quality images and fail to converge in many settings. An improvement \cite{gulrajani2017improved} was proposed to overcome these problems. Although they argued that the performance is more stable at convergence, WGAN is still outperformed by DCGAN \cite{radford2015unsupervised} in terms of  convergent speed and Inception score. A similar solution was introduced in Loss-Sensitive GAN (LS-GAN) \cite{qi2017loss} with theoretical analysis on Lipschitz densities. They conceptually proved that the GAN loss functions with bounded Lipschitz constants are sufficient to match the model density to true data density. However, objects in their generated CIFAR-10 images are hardly recognizable. Meanwhile, Least Squares GAN (LSGAN) \cite{mao2017least} adopted the least square loss function in the discriminator. They showed that minimizing the objective function yields minimizing the Pearson $\mathcal{X}^2$ divergence. Their results demonstrated that LSGAN is able to synthesize appealing images on LSUN, CIFAR-10, and handwritten Chinese characters datasets.

Recently, another subfamily of GAN was introduced where an autoencoder is employed in the discriminator. The Energy-based GAN (EBGAN) \cite{zhao2016energy} is trained by replacing the discriminator with an autoencoder and it has demonstrated decent quality synthetic images up to $256\times256$ pixels. Denoising Feature Matching (DFM) \cite{warde2017improving} maintains the traditional GAN adversarial loss, but an additional complementary information to the generator is computed using a denoising autoencoder in the feature space learned by the discriminator. DFM achieved state-of-the-art Inception score on CIFAR-10 in the unsupervised settings. Both works suggested a non-trivial idea that the multi-targets information from the reconstruction loss helps to improve the model performance. A closely related work, Boundary Equilibrium GAN (BEGAN) \cite{berthelot2017began} was proposed with a new equilibrium enforcing method. Surprisingly, it demonstrated realistic face generation but is significantly outperformed by DFM on CIFAR-10. This suggests that the traditional adversarial loss remains an important factor to generate realistic complex images. 

StackGAN \cite{zhang2016stackgan} was proposed to overcome the instability issue when training GAN to generate images at higher resolutions (e.g. $256\times256$). It employed a hierarchical structure by stacking multiple generators that learn to generate images with different resolutions. Their results demonstrated that StackGAN is able to generate appealing images at $256\times256$ resolution. A different type of hierarchical structure was employed in Karras \etal \cite{karras2017progressive} by progressively training different layers in a generator at different stages. As a result of this, they are able to generate high quality images with resolution as high as $1024\times1024$.

Among recent works, few GAN variants such as CVAE-GAN \cite{bao2017cvae}, LSGAN \cite{mao2017least}, Stacked GAN (SGAN) \cite{huang2016stacked}, and Progressive GAN \cite{karras2017progressive} demonstrated their ability in generating high quality images. Qualitatively, their generated images seem to outperform the proposed ArtGAN in terms of subjective image quality. Interestingly, the proposed ArtGAN is able to achieve better Inception score when compared to SGAN \cite{huang2016stacked}. This shows that Inception score \cite{salimans2016improved} is unable to measure the perceptual quality of an image. 

\subsection{Conditional Image Synthesis}

While unconditional image synthesis is an important research area, many practical applications require the model to be conditioned on some prior information. This prior information has many forms, for instance a distorted image for inpainting \cite{pathak2016context, oord2016pixel}; natural image for super-resolution \cite{denton2015deep} or style transfer \cite{isola2016image, wang2017bayesian, elad2017style}; text codes for text to image translation \cite{reed2016generative, zhang2016stackgan}. Due to the nature of this work, we will only focus on the works related to class-conditioned image generation. 

An earlier work that employed conditional setting in GAN was Conditional GAN (CondGAN) \cite{mirza2014conditional} where it feeds the labels or modes to the generator and discriminator. However, such setting was demonstrated on less complex images \ie MNIST and faces \cite{gauthier2014conditional}. While this website\footnote{\url{http://soumith.ch/eyescream/}} unofficially generated images on CIFAR-10 using CondGAN, the objects in their generated images are hardly recognizable. This is expected because the labels were not fully utilized, as there is no error information backpropagated from the labels. A closed work to ArtGAN is InfoGAN \cite{chen2016infogan} where the discriminator is replaced by a multi-class classifier. Also, InfoGAN has two heads in the discriminator that output $c$ and $y$ separately. Hence, InfoGAN has different architecture compared to ArtGAN. Empirical results showed that InfoGAN is able to learn disentangled representations in an unsupervised manner but the meaning of the representations are uncontrollable during the training stage. As to CondGAN \cite{mirza2014conditional}, InfoGAN only demonstrated on less complex images, \ie digits and faces. Bao \etal \cite{bao2017cvae} proposed CVAE-GAN that combined Conditional Variational Autoencoder and GAN. CVAE-GAN is asymmetrically trained by introducing a new objective function for the generator. At the same time, they also trained an encoder network to map the real image  to the latent vector. This allows their model to learn a better correlation between the latent vector and the image. They demonstrated that CVAE-GAN is able to generate realistic and diverse images on face, flowers, and birds datasets \cite{LFWTechUpdate, Nilsback08, WahCUB_200_2011}. However, CVAE-GAN was trained on pre-processed images centered around the objects. Hence, their results are not comparable to ArtGAN as the images used in our experiments are randomly cropped.

In addition to the GAN variants, PixelCNN \cite{van2016conditional, salimans2017pixelcnn++} also demonstrated decent results on conditional image generation but it is computationally expensive for sampling. Built on Deep Generator Network (DGN) \cite{nguyen2016synthesizing}, Plug and Play Generative Networks (PPGN) \cite{nguyen2016plug} is able to produce high quality images at high resolution. It allows different generators and condition networks to be hacked together without having to re-train the generators. However, PPGN differs to the other generative models discussed, herein images are generated in \textbf{one-shot} from the latent codes in the traditional generative models. That is to say, in PPGN, images are generated by optimizing the latent codes to produce images that highly activate target neuron in the condition network. The sampling procedure is formalized as an approximate Langevin Markov chain Monte Carlo sampler to ensure diversity. Like other sequential approaches, such gradient-based recursive approach may cause unwanted overhead when deployed in some of the real-world applications, \eg mobile devices. Nonetheless, they showed that adversarial training is crucial to obtain high quality images. 

\begin{figure*}[t]
	\centering
	\includegraphics[width=\textwidth]{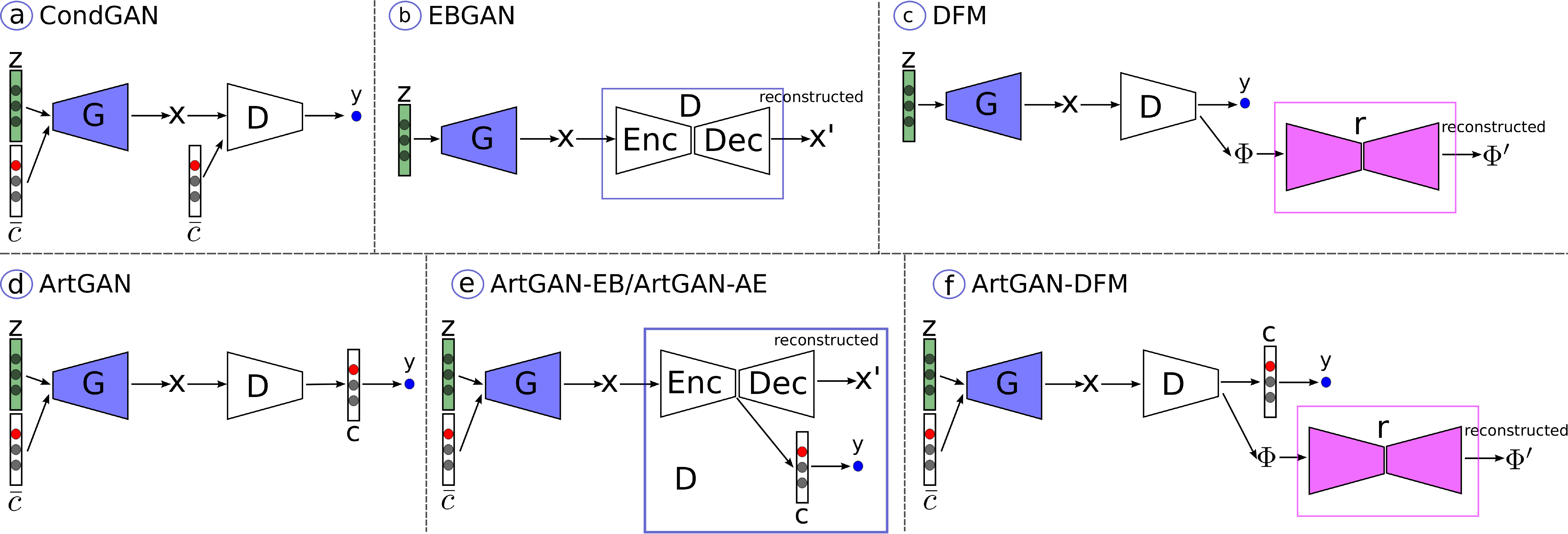}
	\caption{Different ArtGAN variants (bottom row) compared to the state-of-the-art models (top row). The discriminator in ArtGAN outputs the class predictions and the loss function is computed from the true labels, instead of taking the true labels as input as depicted in CondGAN. Hence, the true labels can be leveraged to train the discriminator and generator. Meanwhile, ArtGAN-EB and ArtGAN-AE share the same model, that is a combination of ArtGAN and EBGAN with shared encoder. However, the decoder in ArtGAN-AE is not trained using the generated samples, as opposed to ArtGAN-EB. ArtGAN-DFM depicts the extension from DFM with conditional settings. Note that InfoGAN has a different architecture compared to ArtGAN since InfoGAN has two heads in the discriminator that output class $c$ and adversarial $y$ predictions separately.}\vspace{-.1in}
	\label{artvariant}
\end{figure*}

\section{Proposed Method}
\label{secP}

This section describes the proposed method in detailed. First, we revisit the traditional GAN \cite{goodfellow2014generative} model. Then, we depict the formulations of the proposed ArtGAN variants. The architecture of the best ArtGAN variant (i.e. ArtGAN-AEM) is depicted in Figure \ref{archview}. 

\subsection{Preliminaries: Generative adversarial networks}
\label{secgan}

Generative Adversarial Networks (GAN) \cite{goodfellow2014generative} contains two networks that are trained by competing with each other. The Generator $G$ aims to generate images $G(\mathbf{z})$ that have a distribution $p_G$ similar to the true data distribution $p_{data}$, such that $G(\mathbf{z})$ are difficult to differentiate from real images $\hat{\mathbf{x}}\sim p_{data}$. Traditionally, $G$ generates images from some noise vectors $\mathbf{z}\sim p_{noise}$ that are sampled from a distribution $p_{noise}$ (\eg uniform distribution). On the other hand, the Discriminator $D$ is trained to distinguish the images generated by $G$ from the real images. Overall, the training procedure is a two-player min-max game with the following objective function,
\begin{equation}
\min_G\max_D \mathbb{E}_{\hat{\mathbf{x}}\sim p_{data}}[\log D(\hat{\mathbf{x}})]+\mathbb{E}_{\mathbf{z}\sim p_{noise}}[\log(1-D(G(\mathbf{z}))]
\label{eqgan}
\end{equation} 

\subsection{ArtGAN}
\label{secccgan}
The basic structure of ArtGAN is similar to GAN, such that it consists of a discriminator and a generator that are simultaneously trained using the minmax formulation of GAN, as described in Eq. \ref{eqgan}. The key innovation of ArtGAN is to allow feedback from the labels given to each generated image through the loss function. That is, additional label information is fed to the generator to draw a specific subject based on the information, imitating how human learns to draw. This is in contrast to CondGAN \cite{mirza2014conditional} that does not fully utilize the labels during training. In order to leverage the labels information, the discriminator is extended to \textit{categorical autoencoder-based discriminator} to output $K+1$ logistic predictions with $K$ actual categories following the dataset used, and $K+1^{th}$ output as the adversarial class (denoted as Fake category). 

Formally, the formulation of a categorical discriminator is written as $D:\mathbb{R}^{H\times W\times C}\rightarrow\mathbb{R}^{K+1}$, where $H$, $W$, and $C$ are the height, width, and number of channels of an image, respectively. This is somehow similar to Salimans \etal \cite{salimans2016improved}, except that the conditional setting is not implemented in their work. While the notations of the \textit{conditional generator} is written as $G:(\mathbf{z}, \bar{c})\rightarrow\mathbb{R}^{H\times W\times C}$, where $\bar{c}$ is the randomly chosen label for the generated sample in the form of one-hot vector. This allows the generator to learn better from the feedback labels information. Following Salimans \etal \cite{salimans2016improved}, we modify the categorical discriminator such that $D$ becomes the standard supervised classifier with $K$ outputs, $D:\mathbb{R}^{H\times W\times C}\rightarrow\mathbb{R}^{K}$. Let $l_k(\mathbf{x})\in D(\mathbf{x})$ be the output of $D(\mathbf{x})$ at class $k$ without activation function and $\mathbf{x}$ is an input image (either from real data or generator). The probability distribution over $K$ classes is given as $p(\mathbf{c}|\mathbf{x})$, such that the predicted probability for each class $k$ is defined as a softmax function,

\begin{equation}
p(c_k|\mathbf{x}) = \frac{e^{l_k}}{\sum_{i=1}^K e^{l_i}}
\end{equation}
\noindent The probability distribution function for the binary adversarial prediction $p(y|\mathbf{x})$ of the discriminator is then reformulated as 
\begin{equation}
p(y|\mathbf{x}) = \frac{Z(\mathbf{x})}{Z(\mathbf{x})+1}
\end{equation}
\noindent where $Z(\mathbf{x})=\sum_{i=1}^Ke^{l_i}$. While, $p(y|\mathbf{x})=1$ infers that the image $\mathbf{x}$ is real. The benefit of such setting is that the number of parameters can be reduced to relax the over-parametrization problem without changing the output of the softmax, conceptually. The $D$ is then trained by minimizing the following discriminator loss function $\mathcal{L}_D$,

{\small \begin{align}
\mathcal{L}_D = &- \mathbb{E}_{(\hat{\mathbf{x}},\hat{c})\sim p_{data}}\bigg[\sum_{i=1}^K\hat{c}_i\log p(c_i|\hat{\mathbf{x}}) + \log p(y|\hat{\mathbf{x}})\bigg]  \nonumber \\
&- \mathbb{E}_{\mathbf{z}\sim p_{noise},\bar{c}}\bigg[\log(1-p(y|G(\mathbf{z}, \bar{c})))\bigg]
\label{eqd}
\end{align}}

\noindent where $\hat{c}$ is the ground truth one-hot label of the given real image $\hat{\mathbf{x}}$. The generator loss function $\mathcal{L}_G$ to be minimized for training $G$ is defined as,

{\small \begin{align}
\mathcal{L}_G = &- \mathbb{E}_{\mathbf{z}\sim p_{noise},\bar{c}}\bigg[\sum_{i=1}^K\bar{c}_i\log p(c_i|G(\mathbf{z},\bar{c})) + \log(p(y|G(\mathbf{z}, \bar{c})))\bigg]
\label{eqg}
\end{align}}

Inspired by recent works \cite{warde2017improving, zhao2016energy, berthelot2017began}, we incorporate an autoencoder into the categorical discriminator in ArtGAN for additional complementary information. The core idea of using an autoencoder in the discriminator is that reconstruction-based output offers diverse targets, which produce a very different gradient directions within the minibatch. Conceptually, this improves the efficiency and effectiveness when training a GAN model. Rather than deploying two separate computationally expensive networks (a categorical discriminator and an autoencoder separately), the same architecture and weights are partly shared. In specific, the \textit{encoder} in the autoencoder is shared by the categorical discriminator, as shown in Figure \ref{archview}. In this paper, the formulations of the categorical autoencoder-based discriminators are described in three different ways. The first two variants, ArtGAN-EB and ArtGAN-AE are implemented using the pixel-level autoencoder, similar to EBGAN \cite{zhao2016energy}. However, these two variants are differed in terms of the discriminator loss functions formulation. The third type, ArtGAN-DFM is an extension of Denoising Feature Matching (DFM) \cite{warde2017improving} to a conditional setup, forming a \textit{Conditional DFM}. All the ArtGAN variants are summarized in Figure \ref{artvariant} and the details of the loss functions formulations for each of them are described next. Meanwhile, analysis and comparisons between these ArtGAN variants will be discussed in the experimental section.

\textbf{ArtGAN-EB:} EBGAN \cite{zhao2016energy} is formulated according to the energy-based models by replacing the discriminator with an autoencoder, such that $D_{AE}(\cdot)=Dec(Enc(\cdot))$, where $Dec$ and $Enc$ are the decoder and encoder, respectively. The discriminator loss $\mathcal{L}_{Deb}$ in EBGAN is given as,
 
{\small \begin{align}
\mathcal{L}_{Deb} = &\mathbb{E}_{\hat{\mathbf{x}}\sim p_{data}}\bigg[||D_{AE}(\hat{\mathbf{x}})-\hat{\mathbf{x}}||\bigg] \nonumber \\
&+ \mathbb{E}_{\mathbf{z}\sim p_{noise}}\bigg[\max(0, m - ||D_{AE}(G(\mathbf{z}))-G(\mathbf{z})||)\bigg]
\label{eqdeb}
\end{align}}

\noindent where $||\cdot||$ is a Euclidean norm, and $m$ as a positive margin. The generator loss $\mathcal{L}_{Geb}$ is formulated as, 

\begin{equation}
\mathcal{L}_{Geb} = \mathbb{E}_{\mathbf{z}\sim p_{noise}}\bigg[||D_{AE}(G(\mathbf{z}))-G(\mathbf{z})||\bigg]
\label{eqgeb}
\end{equation}

\noindent In order to formulate a conditional energy-based loss function, ArtGAN-EB propose a novel discriminator loss function $\mathcal{L}_{Debc}$ as,

\begin{equation}
\mathcal{L}_{Debc} = \mathcal{L}_D + \mathcal{L}_{Deb}
\label{eqdebc}
\end{equation}

\noindent and the new generator loss $\mathcal{L}_{Gae}$ is defined as,

\begin{equation}
\mathcal{L}_{Gae} = \mathcal{L}_{G} + \mathcal{L}_{Geb}
\label{eqgae}
\end{equation}

\textbf{ArtGAN-AE:} The discriminator loss is similar to ArtGAN-EB (Eq. \ref{eqdebc}), except that we do not use the generated images as adversarial samples to update the decoder. This was inspired by DFM \cite{warde2017improving} to use the autoencoder as a source of \textit{complementary information} when updating the generator, instead of using the autoencoder as an adversarial function (as in \cite{zhao2016energy}). Hence, the discriminator loss $\mathcal{L}_{Dae}$ of ArtGAN-AE is formulated as,

\begin{equation}
\mathcal{L}_{Dae} = \mathcal{L}_D + \mathbb{E}_{\hat{\mathbf{x}}\sim p_{data}}\bigg[||D_{AE}(\hat{\mathbf{x}})-\hat{\mathbf{x}}||\bigg]
\end{equation}

\noindent Meanwhile, ArtGAN-AE has the same generator loss as ArtGAN-EB (Eq. \ref{eqgae}). 

\textbf{ArtGAN-DFM:} In DFM \cite{warde2017improving}, an additional denoising autoencoder (or denoiser) $r(\cdot)$ is employed to update the generator. The denoiser is trained separately from the discriminator. In specific, the denoiser is trained on the discriminator's hidden state when it is evaluated on the training data. Formally, $D$ is updated according to Eq. \ref{eqgan}. Given that $\Phi(\cdot)$ is a hidden state from $D(\cdot)$, the denoiser is trained by minimizing the following loss function $\mathcal{L}_r$,

\begin{equation}
\mathcal{L}_r = \mathbb{E}_{\hat{\mathbf{x}}\sim p_{data}}\bigg[||\Phi(\hat{\mathbf{x}})-r(\Phi(\hat{\mathbf{x}}))||\bigg]
\label{eqr}
\end{equation}
\noindent Then, the generator is trained with the loss function $\mathcal{L}_{Gdfm}$,
{\small \begin{align}
\mathcal{L}_{Gdfm} = &\mathbb{E}_{\mathbf{z}\sim p_{noise}}\bigg[\lambda_{denoise}||\Phi(G(\mathbf{z}))-r(\Phi(G(\mathbf{z})))|| \nonumber \\
&-\lambda_{adv}\log D(G(\mathbf{z}))\bigg]
\end{align}}

\noindent The authors \cite{warde2017improving} suggested to fix $\lambda_{adv}=1$ and set $\lambda_{denoise}=0.03/n_h$, where $n_h$ is the number of discriminator hidden units fed to the denoiser as input. The modification is straightforward using the categorical discriminator as the discriminator network. Hence, the discriminator loss is same as Eq. \ref{eqd}, and the denoiser loss remains unchanged (Eq. \ref{eqr}). The generator loss $\mathcal{L}_{Gdfmc}$ for the conditional DFM is defined as, 
\begin{equation}
\mathcal{L}_{Gdfmc} = \mathbb{E}_{\mathbf{z}\sim p_{noise}}\bigg[\lambda_{denoise}||\Phi(G(\mathbf{z}))-r(\Phi(G(\mathbf{z})))||\bigg]+\mathcal{L}_G
\end{equation}

\section{Image Quality (IQ) Strategy}
\label{secml}

In order to improve the quality of the image generated by ArtGAN, we introduce a novel strategy. The motivation behind this strategy is to generate a set of pixel to vote for an improved (better quality) pixel via average ranking. That is to say, we will train a generator to synthesize images at a resolution $2\times$ higher than the original image size. Then, these generated images will be downsampled by a factor of 2 via average pooling operation as a voting scheme.

\begin{figure}[t]
\centering
\includegraphics[width=0.7\linewidth]{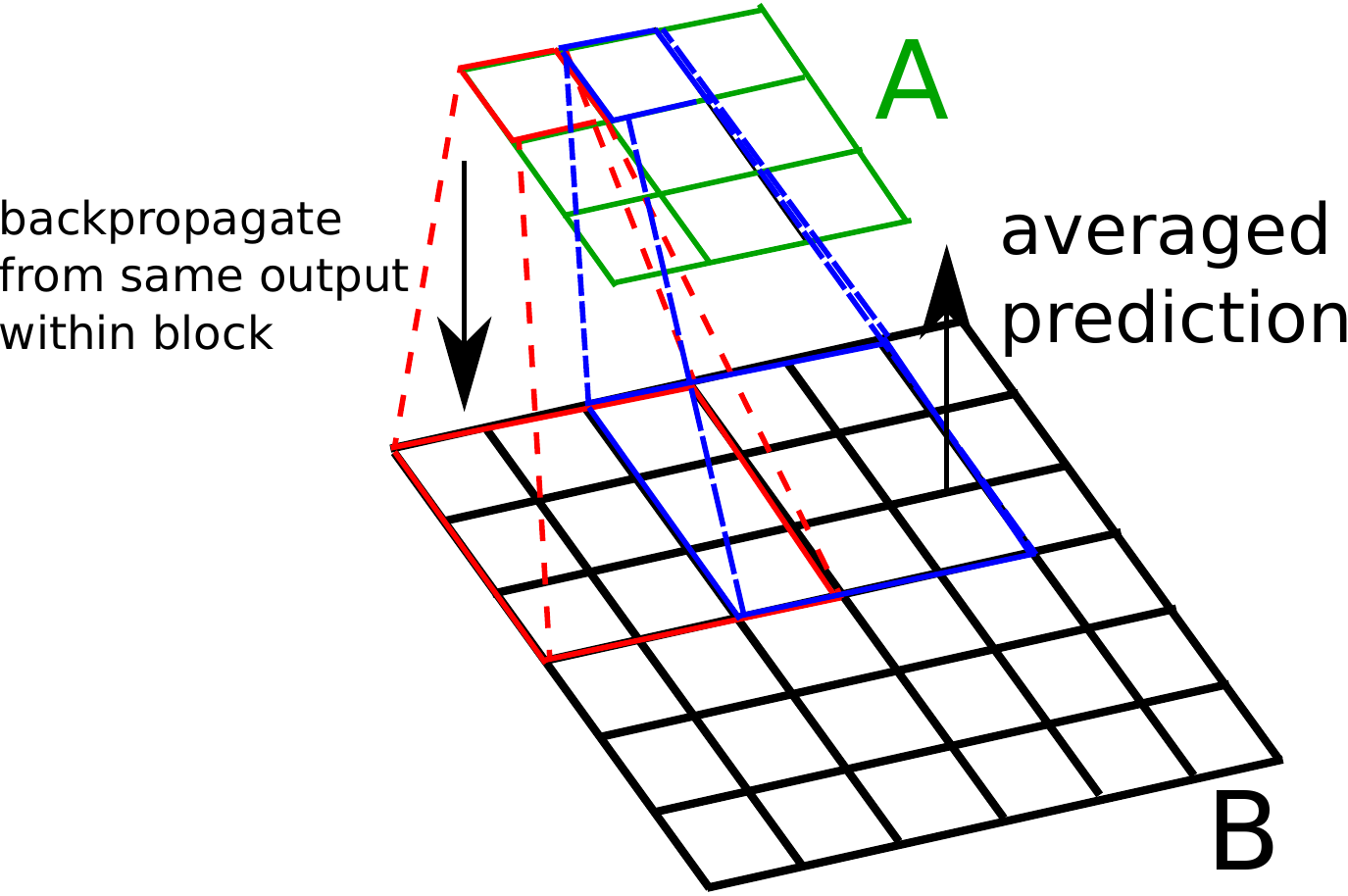}
\caption{The proposed strategy using \textit{overlapped average pooling}. Pixels in the same block from $B$ (\eg $[B_1,\cdots,B_9]$) will vote for an improved pixel in $A$ (\eg $A_1$) through averaging.}\vspace{-.1in}
\label{mlview}
\end{figure}

In specific, suppose a generator in the traditional GAN trained on a dataset generates $32\times32$ pixels images, $G: \mathbf{z}\rightarrow\mathbb{R}^{32\times32\times C}$, where $C$ is the number of channels. Using IQ strategy, the generator will instead generate $64\times64$ pixels images (\ie~$2\times$ higher resolution), $G: \mathbf{z}\rightarrow\mathbb{R}^{64\times64\times C}$. This is done by adding an upsampling block (typically an upsampling layer followed by a convolutinal layer) between the existing layers in the generator. Then, the generated samples are downsampled, such that $\pi:\mathbb{R}^{64\times64\times C}\rightarrow\mathbb{R}^{32\times32\times C}$ where $\pi(\cdot)$ is a downsampling operation. Meanwhile, the input size of the discriminator remains the same as to the original size, such that $D: \mathbb{R}^{32\times32\times C}\rightarrow\mathbb{R}^K$, where $K$ is the number of categories. 

In this paper, overlapped average pooling is chosen as the downsampling operation. The average pooling operation can be viewed as a form of voting system, as shown in Figure \ref{mlview}. Overlapping the pooling operations discourages the generator from blindly computes the same pixel value within the same pooling block. Overall, when the overlapped average pooling is used, the generator is regularized with two seemingly contradictory constraints: i) the generated pixels within the same pooling block should have similar intensity so that the generated image looks smooth across the same color (e.g. smooth blue sky); ii) the generated pixels must not be naively computed to produce the same intensity that may cause excessive artifacts in the image. During inference, this pooling layer can be removed in order to output higher resolution synthetic images. Readers should be noted that this is different from super-resolution as the nature of this paper focuses on generating random images based on the given labels.

\section{Experimental results}
\label{secex}
 
\subsection{Experimental settings}
This section describes the settings that are used in all experiments, unless stated otherwise. All networks are trained with Adam optimizer \cite{kingma2014adam} with an initial learning rate = $0.0002$, $\beta_1 = 0.5$, and minibatch size = 100. The learning rate is decreased by a factor of 10 after iteration $30,000$. Input noise vector $\textbf{z}$ is a 100-dimensional multivariate random variable sampled using an i.i.d. uniform distributed random generator $U(-1, 1)$. Instance noise \cite{sonderby2016amortised} is implemented in all discriminators for better training stability. For a fair comparison, we run one gradient descent step for each player in each iteration. Generally, this is better than running more steps of one player than the other \cite{goodfellow2016nips}. Also in practice, it is very difficult to determine how many steps to use, as the performance is usually inconsistent using the same setting on different datasets. The rest of the settings will be described in the related sections. The experiments were conducted using Tensorflow \cite{abadi2016tensorflow} with one Titan X (Maxwell) GPU. 

For evaluation, Inception score is adopted \cite{salimans2016improved} as the quantitative metric. Intuitively, Inception score measures the \textit{objectness} by minimizing entropy per-sample posterior (\ie each sample is classified with high certainty), as well as the \textit{class diversity} by maximizing the entropy aggregate posterior (\ie the classifier used in Inception score identifies a wide variety of classes among the samples). However, one should aware that \textit{class diversity} metric becomes meaningless in the conditional setting as the conditional generative models will always generate visually different images in different modes. In addition, the \textit{class diversity} metric can be misleading, \ie it can be maximized (higher is better) and fooled when the predicted class distributions of all generated samples are uniform. Hence, we split the measurements (\textit{objectness} and \textit{class diversity} metrics) when we report the scores in this paper for performance evaluations.

Since Inception score is calculated by measuring the object class confidence scores, therefore it is not suitable to assess the model performance on artworks. Meanwhile, evaluation of generative model based on the state-of-the-art log-likelihood estimates can be misleading \cite{theis2015note}. Hence, the comparative studies are first conducted using the \textit{objectness} metric from Inception score on CIFAR-10 \cite{krizhevsky2009learning} and STL-10 \cite{coates2011analysis} datasets. Then, Wikiart dataset \cite{saleh2015large, tan2016ceci} is used for artworks synthesis based on genres, artists, and styles. Finally, we trained on Oxford-102 \cite{Nilsback08} and CUB-200 \cite{WahCUB_200_2011} for additional performance assessments.

We used similar design to BEGAN \cite{berthelot2017began} (i.e. employing nearest neighbour upsampling instead of strided deconvolution layer in the generator as suggested by Odena \etal \cite{odena2016deconvolution}) in order to avoid checkerboard artifacts. Between the upsampling layers, there is at least one layer of convolutional layer. The discriminator has the same design as to the traditional GAN with multiple layers of strided convolutional layers. Batch normalization and leaky ReLU are used in both the discriminator and generator. Due to page limit, detailed network descriptions and additional generated samples are available in the appendix. The list of proposed models are as follows:

\begin{enumerate}
\item ArtGAN - Baseline model \cite{tan2017artgan}. 
\item ArtGAN-EB - The first variant of categorical autoencoder-based discriminator.
\item ArtGAN-AE - The second variant of categorical autoencoder-based discriminator.
\item ArtGAN-DFM - The third variant of categorical autoencoder-based discriminator.
\item ArtGAN-M - ArtGAN with \textbf{IQ strategy}.
\item ArtGAN-D - It has similar architecture as to ArtGAN-M but without IQ strategy. This model is used to verify if network size is the main factor that contributes to the performance improvements observed on ArtGAN-M.
\item ArtGAN-AEM - ArtGAN-AE with \textbf{IQ strategy}.  
\item ArtGAN-AEMT - Huang \etal \cite{huang2016stacked} employed a trick by updating more steps for the generator per each discriminator update step. Although it is hard to determine number of steps, their setup seems to work well for CIFAR-10. Hence, the same setting is employed in our CIFAR-10 experiment as a comparison.
\end{enumerate}

\subsection{Evaluation and metric}

Evaluation of a generative model is extremely difficult as it is still not clear how to quantitatively evaluate a generative model. This is due to the difficulty in estimating the intractable log-likelihood in many models \cite{theis2015note}. The most widely used log-likelihood estimator is the Parzen window estimates \cite{parzen1962estimation}. However, Theis \etal \cite{theis2015note} convincingly argued that this estimator can be quite misleading for high-dimensional data. Recently, Salimans \etal \cite{salimans2016improved} proposed Inception score (higher is better) as a different way to assess image quality by using the: 

\begin{align}
I(\{\mathbf{x}\}_1^N) &= \exp(\mathbb{E}[D_{KL}(p(y|\mathbf{x})||p(y))]) \nonumber \\
&\approx \exp(- \mathbb{E}[H(p(y|\mathbf{x}))] + \mathbb{E}[H(\mathbb{E}_{\mathbf{x}}(p(y|\mathbf{x})))])
\end{align}
\noindent where $H(\cdot)$ is the Shannon entropy and $D_{KL}(\cdot)$ is the Kullback–Leibler divergence. As aforementioned, this metric measures the \textit{objectness} in the first term (lower is better) and \textit{class diversity} in the second term (higher is better) of the samples. It can be misleading when the \textit{class diversity} metric is fooled. An example can be seen in our experiments when we compare ArtGAN (baseline) and ArtGAN-EB in Table \ref{scorecifar}. Although ArtGAN-EB performed better than ArtGAN with higher Inception score (ArtGAN-EB = $8.26$ compared to ArtGAN = $8.21$), it has poor \textit{objectness} score (ArtGAN-EB = $33.51$ compared to ArtGAN = $33.24$). It shows that the \textit{class diversity} score in ArtGAN-EB has affected the Inception score. This is misleading because the combination of high \textit{class diversity} score and poor \textit{objectness} score implies that the objects in the generated images are hard to recognize. Nonetheless, Inception score is still a preferred metric due to the lack of a better alternative for quantitative measurement. Hence, this paper adopts Inception score but the performance assessment is done mainly based on the \textit{objectness} score since it is a more reliable metric. 

In addition, the generated images will be illustrated for visual inspection as human evaluation is always more accurate when accessing the image quality, though can be subjective at times. Furthermore, latent space interpolation is done to ``probe" the structure of the latent space $\mathbf{z}$. Qualitatively, the smooth transitions between samples when the latent space is interpolated usually indicates how well the generative models understand the structure of the images. 

\begin{figure*}[t]
\centering
{\captionsetup{justification=centering}
\begin{subfigure}{0.24\linewidth}
  \centering
  \includegraphics[width=\linewidth]{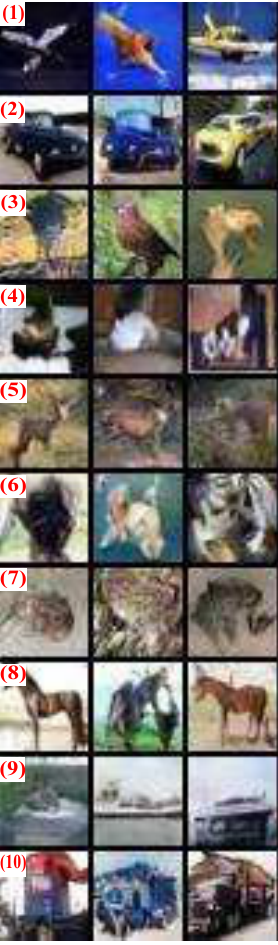}
  \caption{{\footnotesize ArtGAN} ($\mathbf{32\times32}$)}
  \label{cifarccgan}
\end{subfigure}
\begin{subfigure}{0.24\linewidth}
  \centering
  \includegraphics[width=\linewidth]{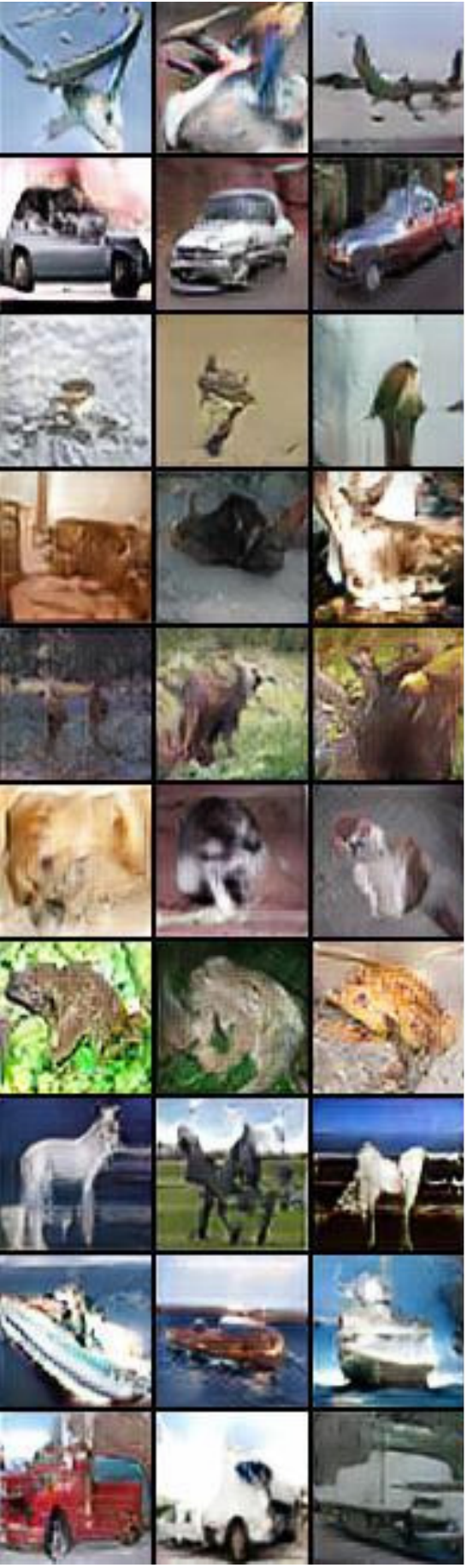}
  \caption{{\footnotesize ArtGAN-M} ($\mathbf{64\times64}$)}
  \label{cifarmccgan}
\end{subfigure}
\begin{subfigure}{0.24\linewidth}
	\centering
	\includegraphics[width=\linewidth]{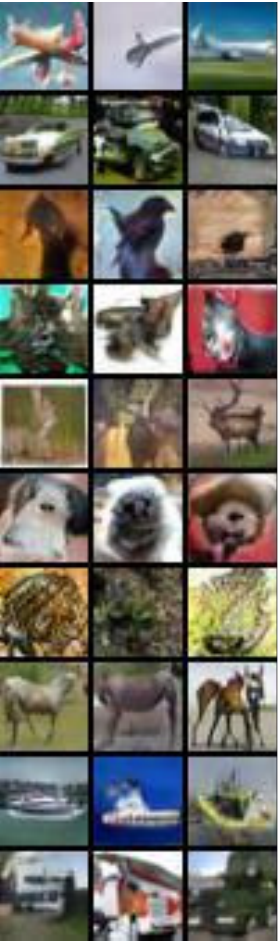}
	\caption{{\footnotesize ArtGAN-AE} ($\mathbf{32\times32}$)}
	\label{cifarccganae}
\end{subfigure}
\begin{subfigure}{0.24\linewidth}
  \centering
  \includegraphics[width=\linewidth]{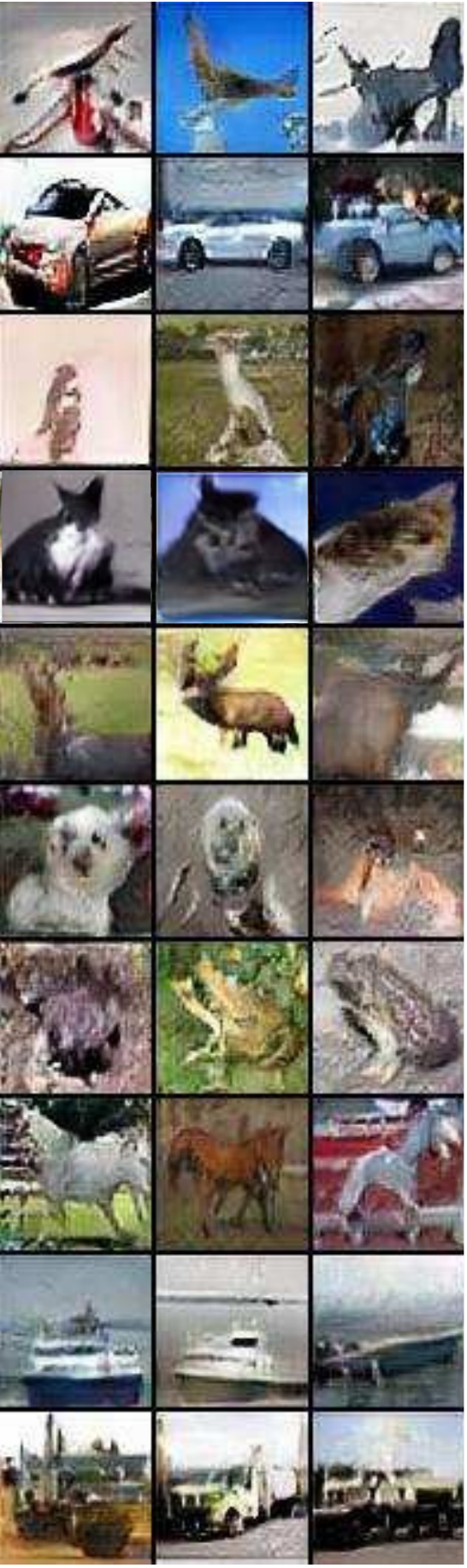}
  \caption{{\scriptsize ArtGAN-AEM} ($\mathbf{64\times64}$)}
  \label{cifarmacan}
\end{subfigure}}
\caption{Comparison of generated CIFAR-10 images with (\ie ArtGAN-M \& ArtGAN-AEM) / without (\ie ArtGAN \& ArtGAN-AE) IQ strategy. Image class from top to bottom: (1) Airplane, (2) Automobile, (3) Bird, (4) Cat, (5) Deer, (6) Dog, (7) Frog, (8) Horse, (9) Ship, (10) Truck. (Best viewed in colour)}
\label{cifars}
\end{figure*}

\begin{table}[t]
	\centering
	\caption{Inception scores on CIFAR-10 evaluated at $32\times32$ pixels. Scores are reported in the form of \textit{mean score}$\pm$\textit{std.} In the proposed methods column, the italic score is objectness metric reported in the form of \textit{objectness} (\textit{class diversity}).}
	\label{scorecifar}
	\begin{tabular}{c|c}
		Model & Scores  \\
		\hline
		\hline
		\multicolumn{1}{l|}{\textit{Unlabelled}} & \\
		\hline
		\hline
		Infusion training \cite{bordes2017learning} & $4.62\pm0.06$ \\
		ALI \cite{dumoulin2016adversarially} (as reported in \cite{warde2017improving}) & $5.34\pm0.05$ \\
		BEGAN \cite{berthelot2017began} & $5.62$ \\
		GMAN \cite{durugkar2016generative} & $6.00\pm0.19$ \\
		EGAN-Ent-VI \cite{dai2017calibrating} & $7.07\pm0.10$ \\
		LR-GAN \cite{yang2017lrgan} & $7.17\pm0.07$ \\
		Denoising feature matching \cite{warde2017improving} & $7.72\pm0.13$ \\
		\hline
		\hline
		\multicolumn{1}{l|}{\textit{Labelled}} & \\
		\hline
		\hline
		SteinGAN \cite{wang2016learning} & $6.35$ \\
		DCGAN (as reported \cite{wang2016learning}) & $6.58$ \\
		Improved GAN \cite{salimans2016improved} & $8.09\pm0.07$ \\
		AC-GAN \cite{odena2016conditional} & $8.25\pm0.07$ \\
		SGAN \cite{huang2016stacked} & $8.59\pm0.12$ \\
		\hline
		\hline
		\multicolumn{1}{l|}{\textit{Proposed methods}} & \\
		\hline
		\hline
		\multirow{2}{*}{ArtGAN (baseline)} & $8.21\pm0.08$ \\
		& \textit{33.24} (\textit{272.90}) \\
		\hdashline
		\multirow{2}{*}{ArtGAN-EB} & $8.26\pm0.10$ \\
		& \textit{33.51} (\textit{276.60}) \\
		\hdashline
		\multirow{2}{*}{ArtGAN-AE} & $8.43\pm0.09$ \\
		& \textit{31.09} (\textit{262.04}) \\
		\hdashline
		\multirow{2}{*}{ArtGAN-DFM} & $8.25\pm0.09$ \\
		& \textit{33.34} (\textit{274.99}) \\
		\hdashline
		\multirow{2}{*}{ArtGAN-M} & $8.50\pm0.06$ \\
		& \textit{30.19} (\textit{256.62}) \\
		\hdashline
		\multirow{2}{*}{ArtGAN-D} & $8.29\pm0.10$ \\
		& \textit{33.30} (\textit{276.15}) \\
		\hdashline
		\multirow{2}{*}{\textbf{ArtGAN-AEM}} & $8.53\pm0.09$ \\
		& \textbf{\textit{30.07}} (\textit{256.42}) \\
		\hdashline
		\multirow{2}{*}{\textbf{ArtGAN-AEMT}} & $\mathbf{8.81}\pm0.14$ \\
		& \textit{30.65} (\textit{269.83}) \\
		\hline
		\hline
		\multirow{2}{*}{Real data} & $11.24\pm0.12$ \\
		& \textit{24.32} (\textit{271.76})
	\end{tabular}
\end{table}

\subsection{CIFAR-10}
CIFAR-10 \cite{krizhevsky2009learning} is a small, well-studied dataset consisting $32\times32$ pixels RGB images. It is split into 50,000 training images and 10,000 test images from 10 classes: airplane, automobile, bird, cat, deer, dog, frog, horse, ship, and truck. 

All ArtGAN variants are trained on full image size, \ie $32\times32$ pixels. When IQ strategy is employed, the generator is able to generate images at a higher resolution (i.e. $64\times64$). All models are trained for 70,000 iterations and saved every 1,000 iterations. As stated by Gulrajani \etal \cite{gulrajani2017improved}, Inception scores of the generative models will continue to oscillate with non-negligible amplitude at convergence. Hence, only the best models found based on the \textit{objectness} score are reported in Table \ref{scorecifar} along with the state-of-the-art results. ArtGAN-AEMT obtains state-of-the-art result with a score of ${8.81}\pm0.14$, outperformed two latest methods - SGAN \cite{huang2016stacked} (${8.59}\pm0.12$) and AC-GAN \cite{odena2016conditional} (${8.25}\pm0.07$). Qualitatively, the proposed models also able to produce many samples with high visual fidelity, especially when IQ strategy is employed as shown in Figure \ref{cifars}. Particularly, the samples drawn by ArtGAN-AEM have finer details, \eg cats are more recognizable with better ears shape (row 1 and 2), most of the frogs are drawn with clear contour (row 2 and 3), etc. 

Interestingly, SGAN \cite{huang2016stacked} demonstrated subjectively better image quality despite lower Inception score when compared to the proposed ArtGAN-AEMT. In SGAN \cite{huang2016stacked}, similar loss function is used for their conditional loss, i.e. cross-entropy for labels. Hence, we deduce that network design and training procedure (e.g. training the networks in a hierarchical manner as in SGAN \cite{huang2016stacked} and Progressive GAN \cite{karras2017progressive}) are important factors for achieving better perceptual image quality. Meanwhile, the proposed ArtGAN baseline has lower subjective image quality and Inception score ($8.21\pm0.08$) when compared with SGAN. Hence, it is clear that the proposed IQ strategy helps improve the Inception score but not the image quality. We deduce that the higher feature dimension introduced by IQ strategy complements the loss function by learning richer representations. This encourages the generation of images that are easy to categorize, resulting in higher Inception score. 


\begin{figure*}[t]
	\centering
	{\captionsetup{justification=centering}
		\begin{subfigure}{0.24\linewidth}
			\centering
			\includegraphics[width=\linewidth]{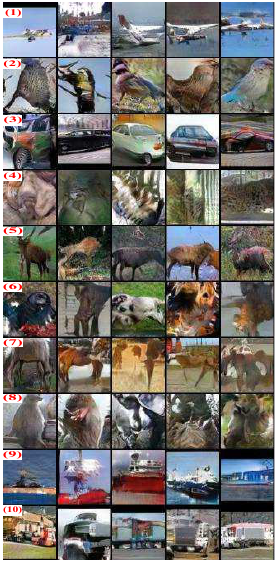}
			\caption{ArtGAN ($\mathbf{64\times64}$)}
			\label{stlccgan}
		\end{subfigure}
		\begin{subfigure}{0.24\linewidth}
			\centering
			\includegraphics[width=\linewidth]{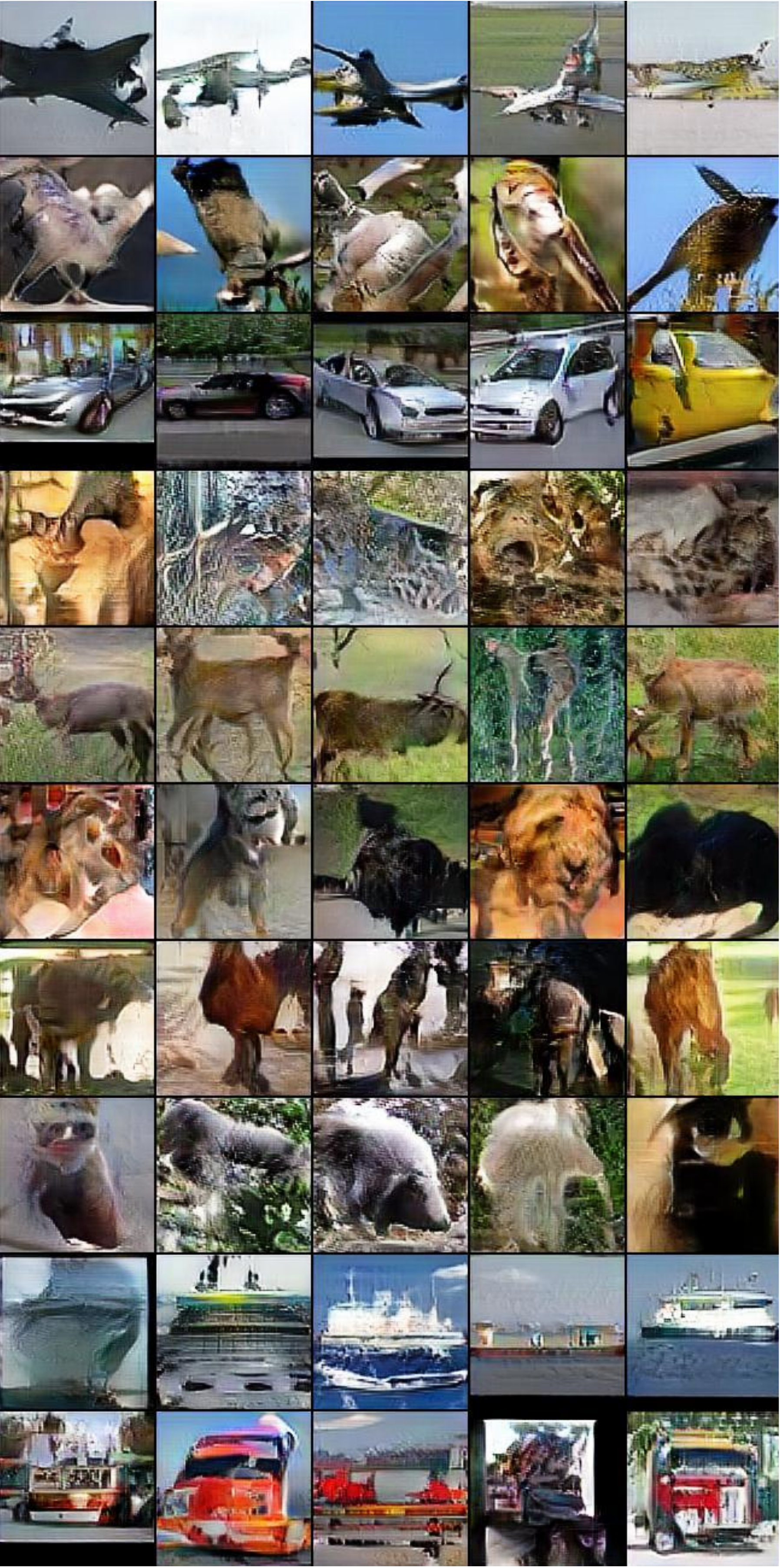}
			\caption{ArtGAN-M ($\mathbf{128\times128}$)}
			\label{stlmccgan}
		\end{subfigure}
		\begin{subfigure}{0.24\linewidth}
			\centering
			\includegraphics[width=\linewidth]{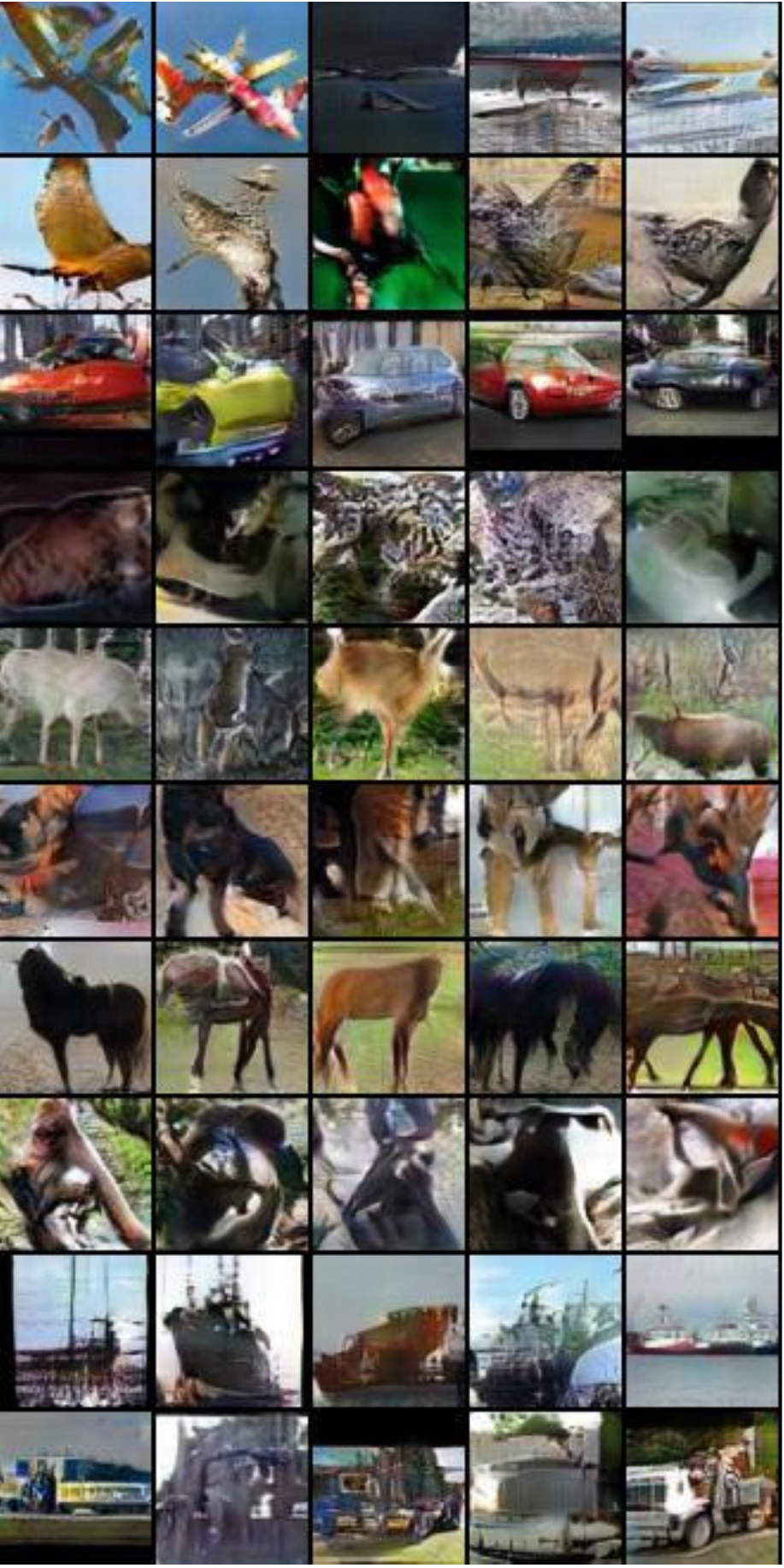}
			\caption{ArtGAN-AE ($\mathbf{64\times64}$)}
			\label{stlccganae}
		\end{subfigure}
		\begin{subfigure}{0.24\linewidth}
			\centering
			\includegraphics[width=\linewidth]{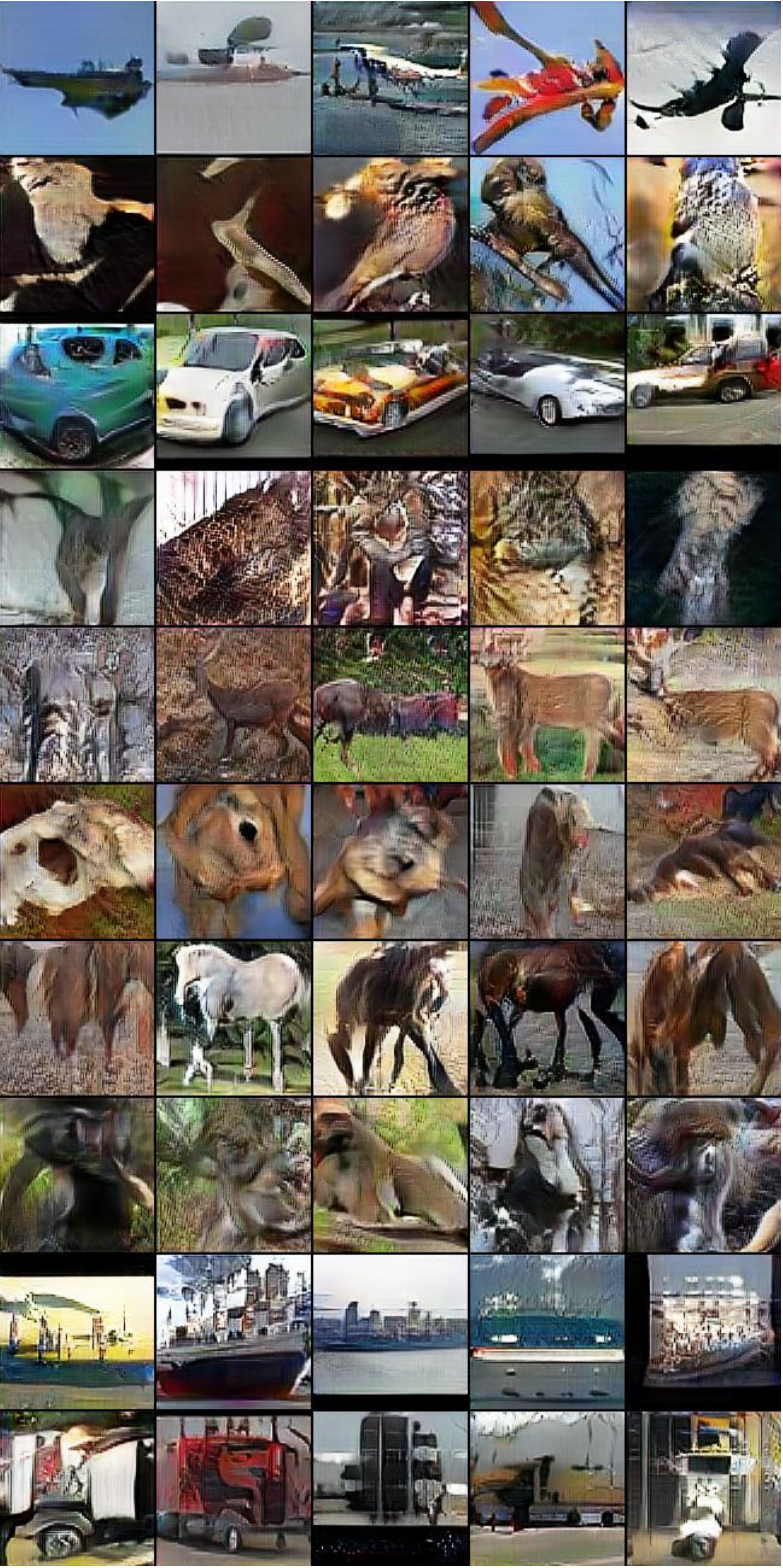}
			\caption{ArtGAN-AEM ($\mathbf{128\times128}$)}
			\label{stlmacan}
	\end{subfigure}}
	\caption{Comparison of generated STL-10 images with (\ie ArtGAN-M \& ArtGAN-AEM) / without (\ie ArtGAN \& ArtGAN-AE) IQ strategy. Image class from top to bottom: (1) Airplane, (2) Bird, (3) Car, (4) Cat, (5) Deer, (6) Dog, (7) Horse, (8) Monkey, (9) Ship, (10) Truck. (Best viewed in colour)} 
	\label{stls}
\end{figure*}

\subsection{STL-10}
STL-10 \cite{coates2011analysis} is a dataset inspired by CIFAR-10 with higher image resolution (\ie~$96\times96$ pixels). However, it contains fewer labelled training examples and has a very large set of unlabelled examples. Although STL-10 is primarily used for unsupervised learning, we employed the dataset for conditional image synthesis in a supervised fashion. In particular, we only employed the labelled examples during training, which contains 5,000 samples from 10 classes: airplane, bird, car, cat, deer, dog, horse, monkey, ship, and truck. As such, it makes STL-10 a more challenging dataset than CIFAR-10. 

During training, we randomly cropped $84\times84$ pixels from the $96\times96$ pixels images. Then, the images are resized and trained at $64\times64$ resolution. Meanwhile, the proposed models trained using IQ strategy are able to generate samples at $128\times128$ resolution. All models are trained for 50,000 iterations. Similar to CIFAR-10, models are saved every 1,000 iterations and the scores of the best models are reported. The Inception scores are reported in Table \ref{scorestl}, while the generated samples are shown in Figure \ref{stls}. It can be noticed that the synthetic images generated by ArtGAN-AEM trained with IQ strategy are clearer and sharper without much artifacts. For instance, the face features of the dogs are more recognizable (row 1-3). No mode collapse is observed in this experiment. 

\begin{table}[t]
\centering
\caption{Inception scores on STL-10 evaluated at $64\times64$ pixels. Readers may refer to Table \ref{scorecifar} for scores descriptions.}
\label{scorestl}
\begin{tabular}{c|c}
Model & Scores  \\
\hline
\multirow{2}{*}{ArtGAN (baseline)} & $9.72\pm0.14$ \\
& \textit{31.03} (\textit{301.63}) \\
\hdashline
\multirow{2}{*}{ArtGAN-EB} & $9.73\pm0.12$ \\
& \textit{30.22} (\textit{293.89}) \\
\hdashline
\multirow{2}{*}{ArtGAN-AE} & $9.65\pm0.08$ \\
& \textit{31.04} (\textit{299.50}) \\
\hdashline
\multirow{2}{*}{ArtGAN-DFM} & $9.63\pm0.09$ \\
& \textit{31.25} (\textit{300.89}) \\
\hdashline
\multirow{2}{*}{ArtGAN-M} & $\mathbf{10.12}\pm0.09$ \\
& \textit{29.05} (\textit{293.90}) \\
\hdashline
\multirow{2}{*}{ArtGAN-D} & $9.87\pm0.09$ \\
& \textit{31.03} (\textit{306.39}) \\
\hdashline
\multirow{2}{*}{\textbf{ArtGAN-AEM}} & $10.07\pm0.09$ \\
& $\mathbf{\textbf{28.18}} (\textit{283.81})$ \\
\hline
\multirow{2}{*}{Real data} & $15.48\pm0.76$ \\
& \textit{15.04} (\textit{232.17}) 
\end{tabular}
\end{table}

\subsection{More ablation studies}
In order to further understand the effects of different ArtGAN variants, we conduct extensive ablation studies by comparing the performances of the ArtGAN models on CIFAR-10 (Table \ref{scorecifar}) and STL-10 (Table \ref{scorestl}) datasets. Note that the performances are evaluated based on the \textit{objectness} metric only, unless specified otherwise. Below we summarize our findings.
   
First, the effectiveness of the IQ strategy can be assessed by comparing ArtGAN-M with the baseline (ArtGAN) and ArtGAN-D. Although ArtGAN-D has more parameters than the baseline, it does not exhibit overfitting problem since its performance is similar to the baseline. We can notice that ArtGAN-M outperformed the baseline and ArtGAN-D significantly. This shows that the extra convolutional layers in the generator are not the main factor that contribute to the performance improvement when the IQ strategy is employed. This is because both ArtGAN-M and ArtGAN-D have the same number of layers, so it proves that the additional upsampling layer introduced in the IQ strategy is the main reason for the improvement. We deduce that richer representation can be learned with higher feature dimension.

Second, ArtGAN-DFM performed poorer than the baseline. In ArtGAN-DFM, the features fed to the denoiser are extracted from the discriminator that is still in training mode. Hence, we speculate that measuring the loss using these primitive features might cause instability when training the denoiser and generator. Therefore, we encourage to compute the losses by leveraging the true data directly. 

Third, inconsistent performance can be noticed in ArtGAN-EB, where it performed best on one dataset but worst on the other. This suggests that additional adversarial loss does not always complement a model. This is because the primitive adversarial samples may provide noisy information that hamper the training process. ArtGAN-AE exhibited more consistent performances with either better or comparable scores. We also trained another variant using only the Energy-based adversarial loss (\ie~traditional adversarial loss is removed). Unfortunately, we found that this model failed to learn, produced collapsed and meaningless images. This deduces that traditional adversarial loss is still a better choice for adversarial training. 

Finally, ArtGAN-AEM (ArtGAN-AE with IQ strategy) achieved the best results with consistent and significant improvements. Meanwhile, ArtGAN-AEMT has the best overall Inception score on CIFAR-10 (8.81).

\begin{figure}[t]
	\centering
	{\captionsetup{justification=centering}
		\begin{subfigure}{0.45\linewidth}
			\centering
			\includegraphics[width=\linewidth]{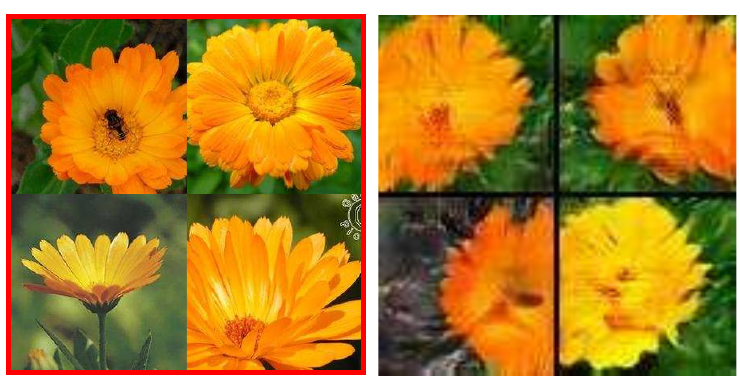}
			\caption{Barbeton Daisy}
			\label{flower1}
		\end{subfigure}
		\begin{subfigure}{0.45\linewidth}
			\centering
			\includegraphics[width=\linewidth]{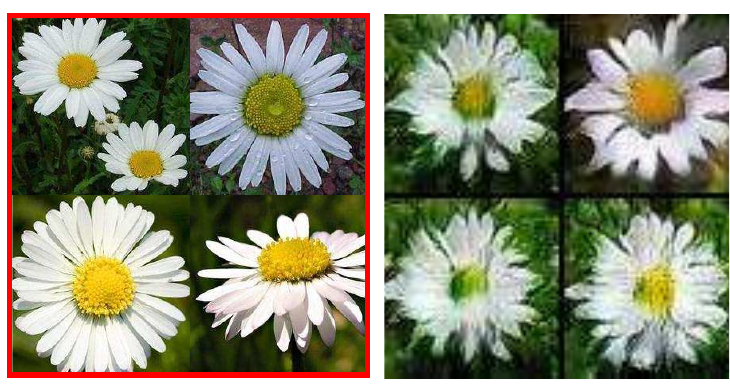}
			\caption{Oxeye Daisy}
			\label{flower2}
		\end{subfigure}
		\begin{subfigure}{0.45\linewidth}
			\centering
			\includegraphics[width=\linewidth]{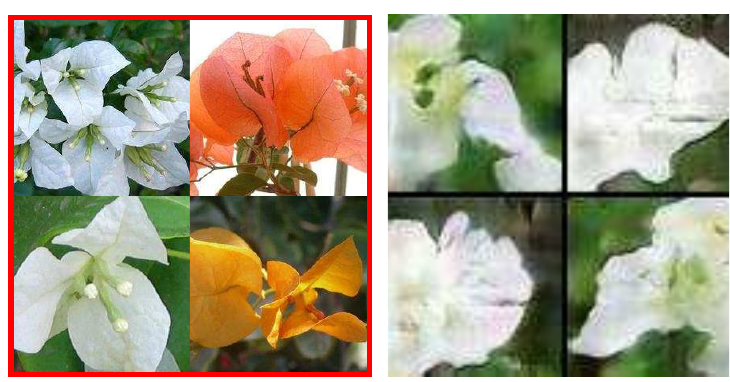}
			\caption{Globe Thistle}
			\label{flower3}
		\end{subfigure}
		\begin{subfigure}{0.45\linewidth}
			\centering
			\includegraphics[width=\linewidth]{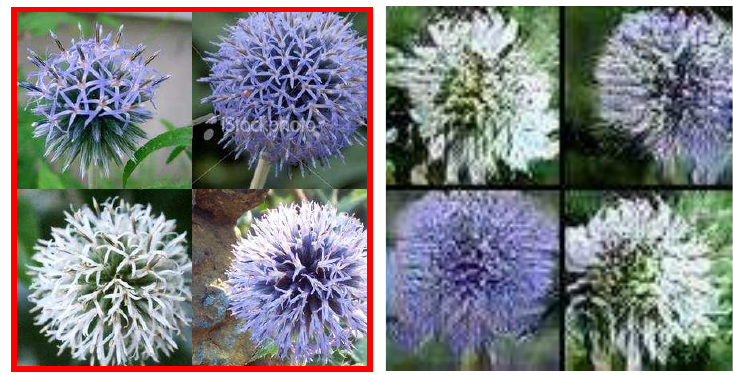}
			\caption{Bougainvillea}
			\label{flower4}
	\end{subfigure}}
	\caption{Sample generated images on Oxford-102 flowers. Left (red box): Real samples; Right: Generated samples. (Best viewed in colour)} 
	\label{flowers}
\end{figure}

\subsection{Oxford-102}
Oxford-102 \cite{Nilsback08} consists of 102 flower species. Each category has around 40 to 258 samples. The samples have large variations in terms of scale, pose, and light. Beside this, some categories exhibit very similar appearance to each other. The model was trained for 30,000 iterations with learning rate reduced after iteration 15,000. The images were saved at $256\times256$ resolution. During training, the images are randomly cropped at $224\times224$, and then resized to $64\times64$. 

Two experiments were conducted. In the first experiment, batch size = 102 is used. In the generator, one sample is drawn for each class during the training stage. We found out that the image quality is high but it suffered from mode collapse, \ie the generated images look almost exactly the same within a class. In the second experiment, 20 classes are randomly chosen in each iteration and with this, 5 samples are drawn for each class during the training stage. This solved the mode collapse problem, suggests that more adversarial images should be sampled for each class in the same iteration to learn more diverse correlations between the latent codes and the image space. Sample of the generated images are depicted in Figure \ref{flowers}. Although the discriminator performed poorly on the classification of flower species ($\sim 50\%$ accuracy), Figure \ref{flowers} shows that ArtGAN-AEM is able to generate high quality flower images that look natural with distinctive species-typical features, \ie color and shape.

\subsection{CUB-200}
Caltech-UCSD Birds-200-2011 (CUB-200) \cite{WahCUB_200_2011} contains 11,788 samples from 200 bird species. The images are pre-processed in the similar way as to Oxford-102, \ie model is trained at $64\times64$ resolution after cropping and resizing. 

\begin{figure}[t]
	\centering
	{\captionsetup{justification=centering}
		\begin{subfigure}{0.45\linewidth}
			\centering
			\includegraphics[width=\linewidth]{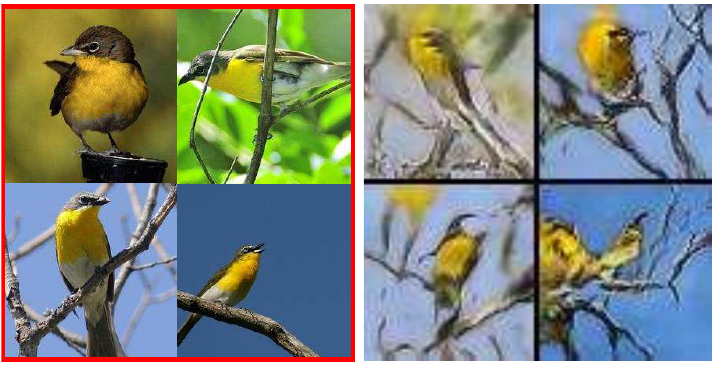}
			\caption{Yellow Breasted Chat}
			\label{bird1}
		\end{subfigure}
		\begin{subfigure}{0.45\linewidth}
			\centering
			\includegraphics[width=\linewidth]{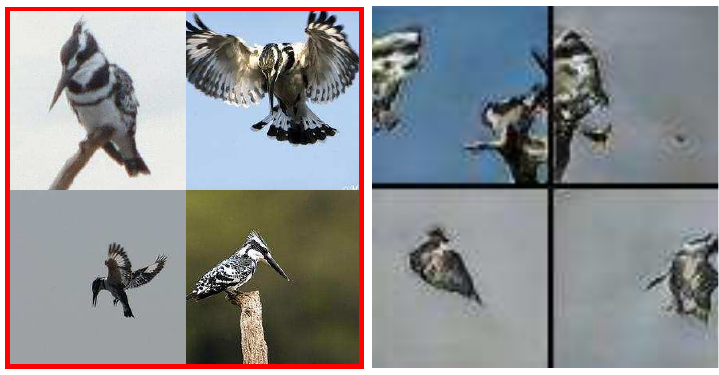}
			\caption{Pled Kingfisher}
			\label{bird2}
		\end{subfigure}
		\begin{subfigure}{0.45\linewidth}
			\centering
			\includegraphics[width=\linewidth]{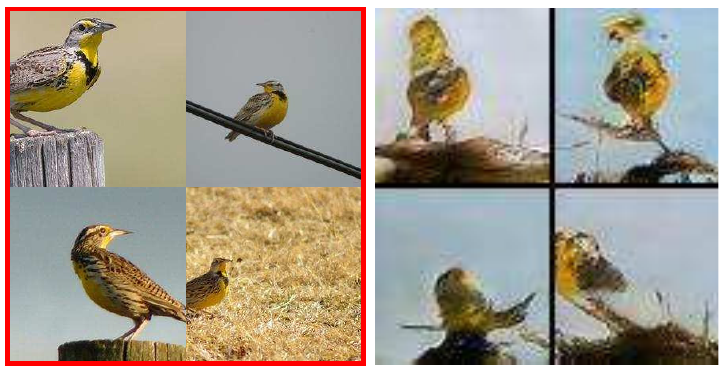}
			\caption{Pied Billed Grebe}
			\label{bird3}
		\end{subfigure}
		\begin{subfigure}{0.45\linewidth}
			\centering
			\includegraphics[width=\linewidth]{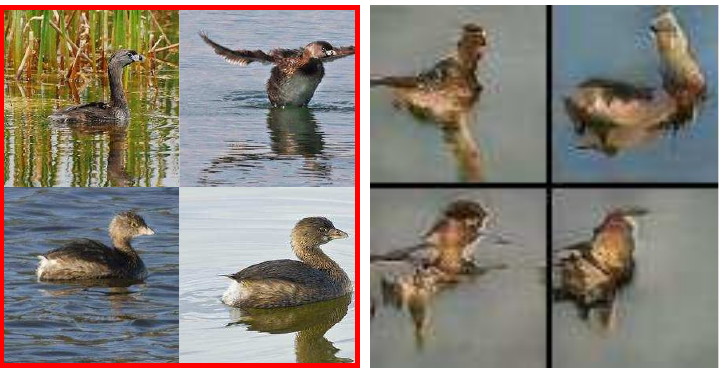}
			\caption{Western Meadowlark}
			\label{bird4}
	\end{subfigure}}
	\caption{Sample generated images on CUB-200 birds. Left (red box): Real samples; Right: Generated samples. (Best viewed in colour)}
	\label{birds}
\end{figure}

In order to avoid the mode collapse experienced in Oxford-102, the model herein follows the same settings (\ie~randomly choosing 20 classes in each iteration with 5 samples per class). The generated image samples are shown in Figure \ref{birds}. Similar to Oxford-102 dataset, the discriminator has a poor performance on the bird species classification ($\sim 20\%$ accuracy). Interestingly, the figures show that ArtGAN-AEM is still able to draw the characteristics of different bird species, \eg colors, shape, and body size. However, the body structures of the birds are not well-learned.

\subsection{WikiArt}
Wikiart is a fine-art paintings dataset first introduced by Saleh \etal \cite{saleh2015large}. The paintings were obtained from the wikiart.org website. Currently, Wikiart is the largest public dataset available that contains around 80,000 annotated paintings for genres, artists and styles classification tasks. However, not all paintings are used in all tasks. To be specific, all paintings are used for 27 \textit{styles} classification. But, there are only 60,000 paintings annotated for 10 \textit{genres}, and only around 20,000 paintings are annotated for 23 \textit{artists}. In this paper, we used an extended version of Wikiart dataset. The extended dataset is randomly split to training and test sets for a fair comparison.

\begin{figure*}[t]
\centering
  \includegraphics[width=.98\linewidth]{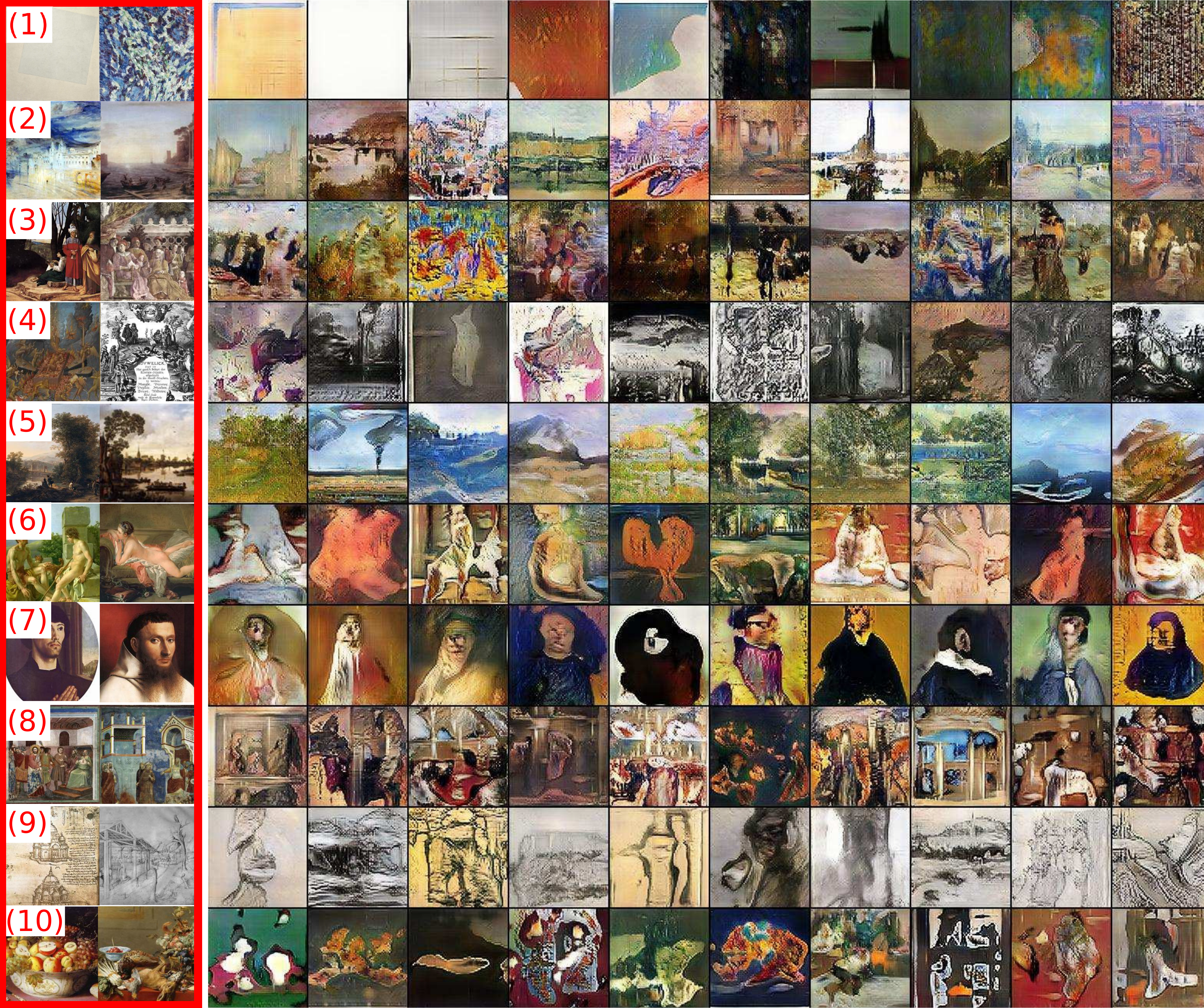}
\caption{Generated \textit{genres} images at $\mathbf{128\times128}$ pixels. Images in the red bounding box are real samples. Genres class from top to bottom: (1) Abstract, (2) Cityscape, (3) Genre painting, (4) Illustration, (5) Landscape, (6) Nude, (7) Portrait, (8) Religious, (9) Sketch and study, (10) Still life. (Best viewed in colour)}
\label{genres}
\end{figure*}

\begin{figure*}[htp]
\centering
  \includegraphics[width=.98\linewidth]{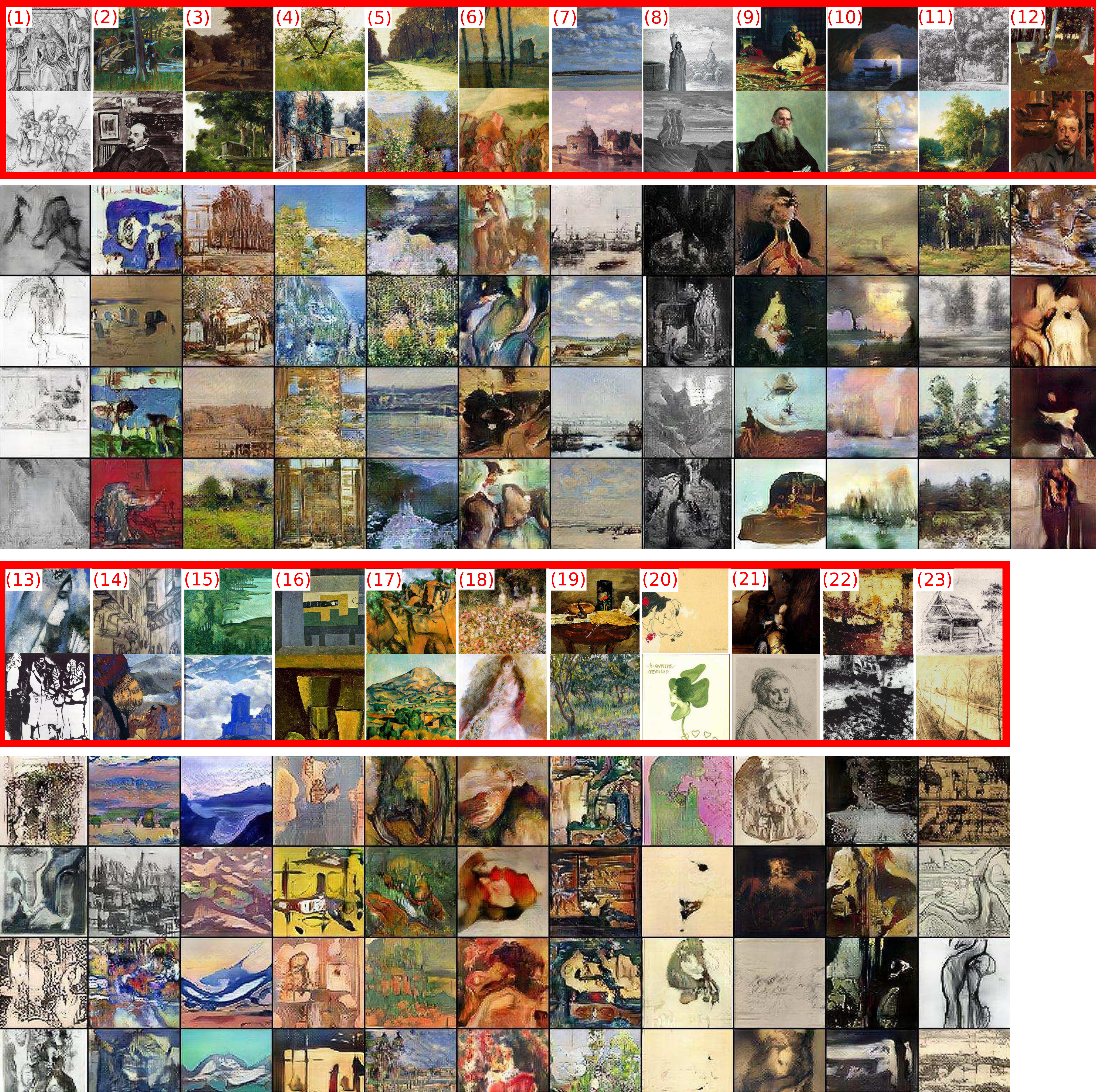}
\caption{Generated \textit{artists} images at $\mathbf{128\times128}$ pixels. Images in the red bounding boxes are real samples. Artists class from left to right, \textbf{Top}: (1) Albrecht Durer, (2) Boris Kustodiev, (3) Camille Pissarro, (4) Childe Hassam , (5) Claude Monet, (6) Edgar Degas, (7) Eugene Boudin, (8) Gustave Dore, (9) Ilya Repin, (10) Ivan Aivazovsky, (11) Ivan Shishkin, (12) John Singer Sargent; \textbf{Bottom}: (13) Marc Chagall, (14) Martiros Saryan, (15) Nicholas Roerich, (16) Pablo Picasso, (17) Paul Cezanne, (18) Pierre Auguste Renoir, (19) Pyotr Konchalovsky, (20) Raphael Kirchner, (21) Rembrandt, (22) Salvador Dali, (23) Vincent van Gogh. (Best viewed in colour)}
\label{artists}
\end{figure*}

The Wikiart images were prepared in $256\times256$ resolution. In this paper, however, at each iteration of the training stage, we randomly cropped the images into $224\times224$ resolution. Since the proposed models are built on standard GAN, we experienced similar problem found in \cite{zhang2016stackgan}. That is, the proposed models are prone to generate nonsensical images when trained using $128\times128$ or higher resolutions images. As a result of that, we resized the cropped images to $64\times64$ resolution. Three different ArtGAN-AEM models were trained for different tasks (\ie~\textit{styles}, \textit{genres} and \textit{artists}) for 50,000 iterations. The results are reported using the final model (\ie model at iteration 50,000). In general, it is observed that ArtGAN-AEM is able to learn artistic representations and generate high quality paintings. Detailed discussions are as follows:

\subsubsection{\textbf{Genre}}
The generated paintings based on genre are showed in Figure \ref{genres}. Out of the three tasks, genres classification can be considered as the most easiest task \cite{tan2016ceci}. Hence, it is expected that ArtGAN-AEM is able to draw many meaningful paintings based on the genre. For instance, one should be able to differentiate \textit{abstract paintings}, \textit{cityscape}, \textit{landscape}, and \textit{portraits} from other classes easily. The synthesized paintings show that ArtGAN-AEM is able to recognize and draw high quality paintings on these genres. An interesting observation can also be observed in the \textit{genre painting} (\ie~No 3). Not to be confused with ``genre", ``genre paintings" is a pictorial representation of scenes or events from everyday life, such as markets, parties, etc. Hence, a group of people is usually visible in this type of paintings. Figure \ref{genres} shows that ArtGAN-AEM is able to draw several human-like figures in a few synthetic paintings (\ie~column 5, 6 and 10 of No 3). The model may not be able to understand the true meaning of \textit{genre paintings}, but this observation shows that ArtGAN-AEM is able to find certain semantic cues. 

\begin{figure*}[t]
	\centering
	\includegraphics[width=.98\linewidth]{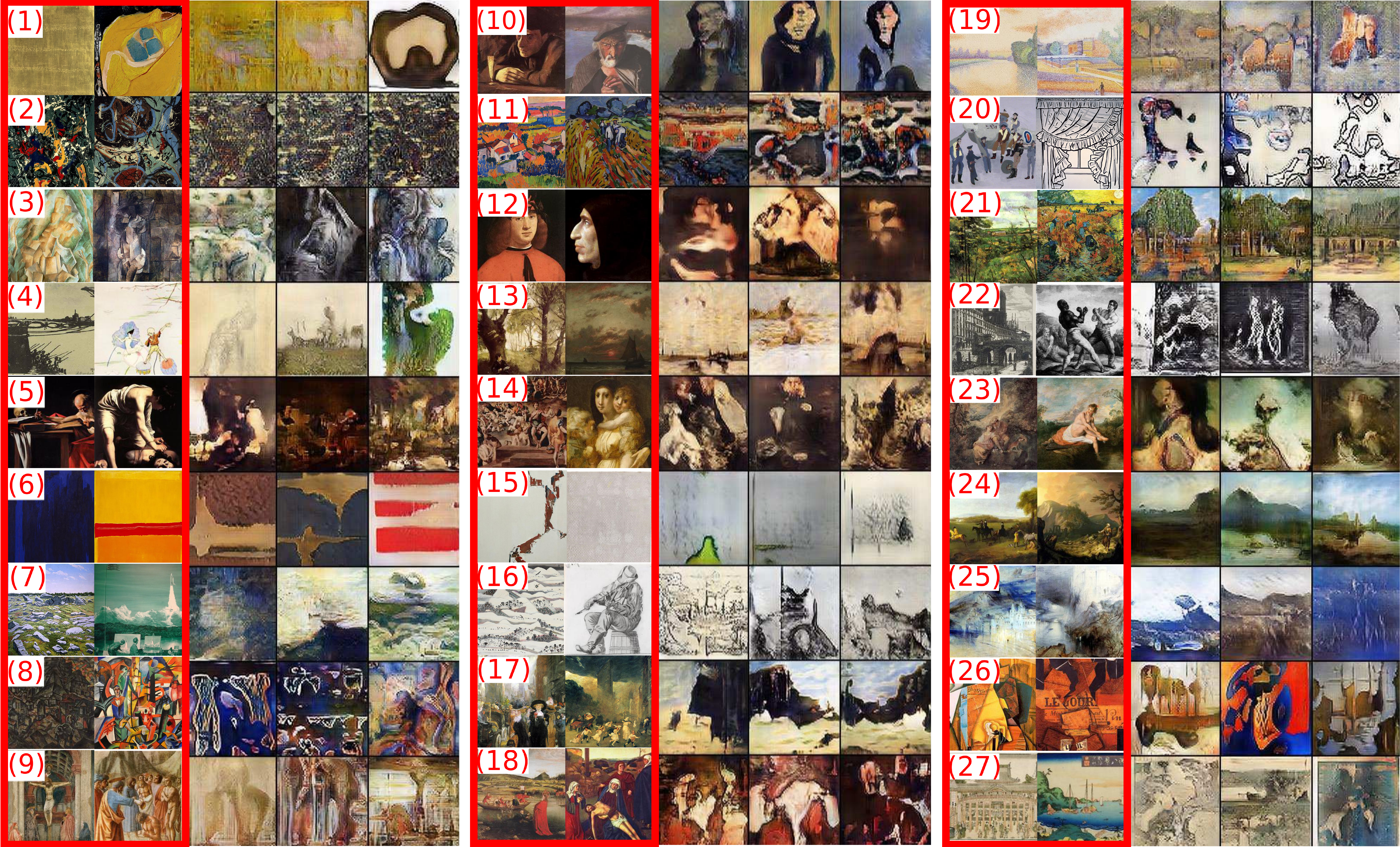}
	\caption{Generated \textit{styles} images at $\mathbf{128\times128}$ pixels. Images in the red bounding boxes are real samples. Styles class from top to bottom, \textbf{Left}: (1) Abstract Expressionism, (2) Action painting, (3) Analytical Cubism, (4) Art Nouveau, (5) Baroque, (6) Color Field Painting, (7) Contemporary Realism, (8) Cubism, (9) Early Renaissance; \textbf{Middle}: (10) Expressionism, (11) Fauvism, (12) High Renaissance, (13) Impressionism, (14) Mannerism Late Renaissance. (15) Minimalism, (16) Naive Art Primitivism, (17) New Realism, (18) Northern Renaissance; \textbf{Right}: (19) Pointillism, (20) Pop Art, (21) Post Impressionism, (22) Realism, (23) Rococo, (24) Romanticism, (25) Symbolism, (26) Synthetic Cubism, (27) Ukiyo-e. (Best viewed in colour)}
	\label{styles}
\end{figure*}

\subsubsection{\textbf{Artist}}
Figure \ref{artists} shows the synthetic paintings based on artist. Learning visual representations in this task is possible as artists usually have their own preferences when deciding what to draw, what kind of styles to use, etc. Hence, many visual similarities can be found from those artworks within the same artist. For example, this can be seen in the paintings of \textit{Nicholas Roerich}. He is a Russian who settled in Himachal Pradesh, India (a mountainous state) for a long time. As a result of that, many of his famous masterpieces depict the beauty of the mountains with expressive colors and fluid brushwork. These characteristics appear in all the synthesized paintings of \textit{Nicholas Roerich} (\ie~No 15). At the same time, all the synthesized paintings of \textit{Gustave Dore} (\ie~No 8) also clearly display his primary approach in engraving, etching, and lithography, which result in grayish artworks. However, the synthesized paintings conditioned on \textit{Vincent van Gogh} appear to be colourless (\ie~No 23). After some investigations, we found an interesting fact that more than half of his artworks were annotated as \textit{sketch and study} genre in the Wikiart dataset. Among all his artworks, most \textit{Van Gogh}’s palette consisted mainly of sombre earth tones, particularly dark brown, and show no sign of the vivid colours that distinguish from his later work, \eg~the famous The Starry Night masterpiece. This explains the behaviour of the trained model. But, this is still not competent as the striking colour, emphatic brushwork, and the contoured forms of his work that powerfully influenced the Expressionism style in modern art is not well-learned by ArtGAN. \textit{Eugene Boudin} is a marine painter and he has always favoured rendering the sea and along its shores in his artworks. Meanwhile, \textit{Ivan Shishkin} became famous for his forest landscapes. All these preferences can be visualize in all the synthesized paintings of \textit{Eugene Boudin} (\ie~No 7) and \textit{Ivan Shishkin} (\ie~No 11), respectively. 

\subsubsection{\textbf{Style}}
Synthetic paintings based on style are shown in Figure \ref{styles}. Out of the three tasks, style is the most difficult task. For instance, as highlighted in Section \ref{secrelated}, it is hard to recognize \textit{Renaissance} art. Beside that, it is also a very challenging task to differentiate \textit{Baroque} and \textit{Rococo} as they are historically related. Generally, they are differentiated by the ``feelings" they give to their viewers (\ie~curator). \textit{Baroque} art often depicts violence, darkness, and the nudes are more plump compared to the \textit{Rococo} artwork. During mid-1700s, artists gradually moved away from \textit{Baroque} into the modern \textit{Rococo} style. \textit{Rococo} art was often light-hearted, pastoral, and a rosy-tinted view of the world. A subjective observation can be seen in Figure \ref{styles} such that \textit{Baroque} synthetic arts (\ie~No 5) are drawn using darker color than the \textit{Rococo} counterparts (\ie~No 23). The color intensity shows that ArtGAN-AEM has managed to learn some of these characteristics. Meanwhile, \textit{Ukiyo-e} is a type of Japanese art flourished from the $17^{th}$ through $19^{th}$ centuries. It is produced using the woodblock printing for mass production and a large portion of these paintings appear to be yellowish due to the paper material. It is observed that such characteristics are generated in the synthetic Ukiyo-e style paintings (\ie~No 27). 

\begin{figure*}[t]
	\centering
	{\captionsetup{justification=centering}
		\begin{subfigure}{.9\linewidth}
			\centering
			\includegraphics[width=\linewidth]{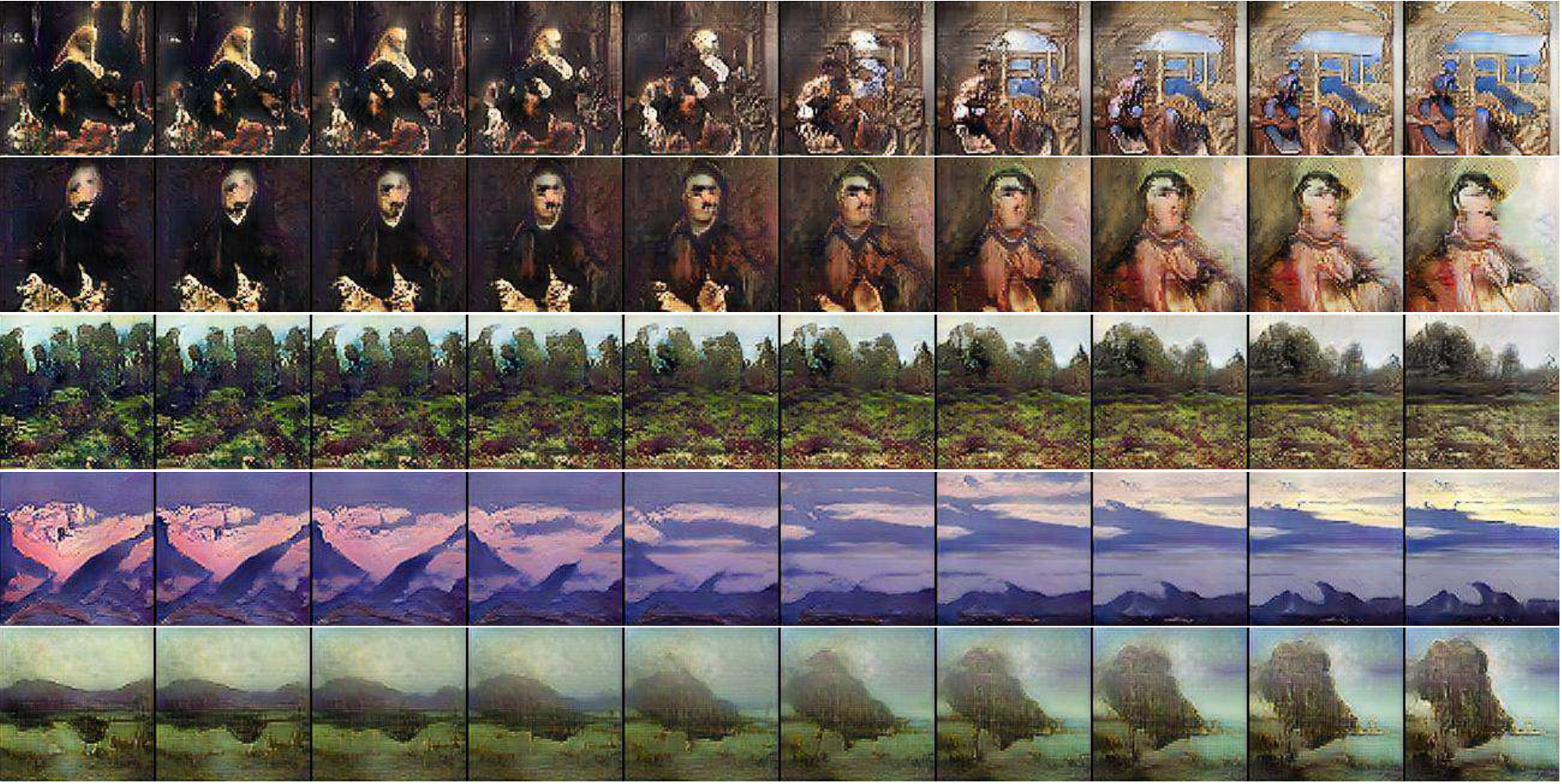}
			\caption{Wikiart}
			\label{inter1}
		\end{subfigure}
		\begin{subfigure}{.9\linewidth}
			\centering
			\includegraphics[width=\linewidth]{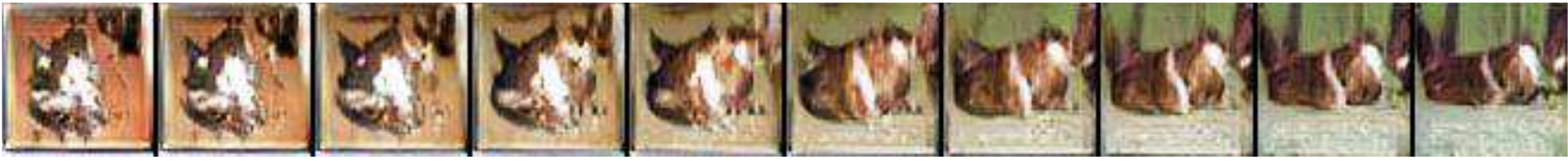}
			\caption{CIFAR-10}
			\label{inter2}
		\end{subfigure}
		\begin{subfigure}{.9\linewidth}
			\centering
			\includegraphics[width=\linewidth]{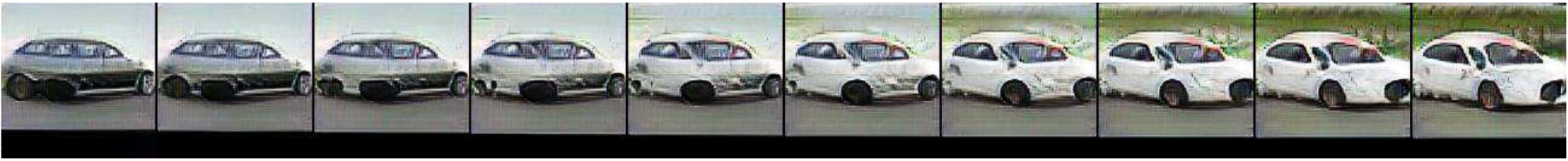}
			\caption{STL-10}
			\label{inter3}
		\end{subfigure}
		\begin{subfigure}{.9\linewidth}
			\centering
			\includegraphics[width=\linewidth]{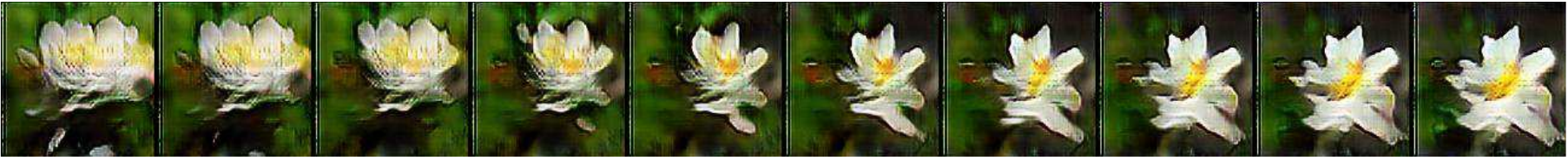}
			\caption{Oxford-10}
			\label{inter4}
		\end{subfigure}
		\begin{subfigure}{.9\linewidth}
			\centering
			\includegraphics[width=\linewidth]{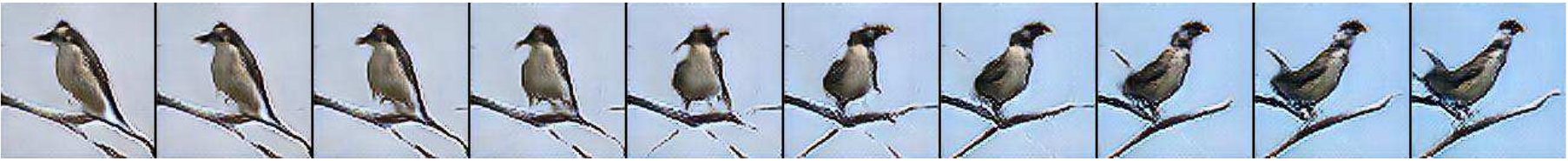}
			\caption{CUB-200}
			\label{inter5}
	\end{subfigure}}
	\caption{Interpolations over the latent space $\mathbf{z}$ in (a) Wikiart, (b) CIFAR-10, (c) STL-10, (d) Oxford-102 and (e) CUB-200 datasets. It demonstrates that ArtGAN does not memorize the training data since samples show smooth transitions and each image looks plausible. (Best viewed in colour)} 
	\label{inter}
\end{figure*}

\subsection{Latent space interpolation}

In this section, we demonstrate that ArtGAN is not simply memorizing the training data, but can truly generate novel images. Walking on the manifold of the latent space $\mathbf{z}$ can examines the signs of memorization, \ie sharp image transitions along the latent space indicate high probability that the model memorizes the true data space. This will be an undesired property as it also implies that the relation between the latent codes and image space is not well learned. Figure \ref{inter} shows that the generated samples have smooth semantic changes and look plausible. For instance, the bird in the synthetic images of CUB-200 rotated from left to right smoothly. This confirms that ArtGAN is not memorizing and has learned relevant, interesting, and rich visual representations. 

\section{Conclusion}
\label{secconclude}

This paper proposed a novel GAN variant called ArtGAN which leverage the labels information for better learning representation and image quality. Empirically, it showed that an extension of ArtGAN (\ie~ArtGAN-AEM) achieved state-of-the-art results on CIFAR-10 and STL-10. Furthermore, ArtGAN-AEM showed the superiority in generating high quality and plausibly looking images on Oxford-102 and CUB-200 datasets. Not to mention, the generated paintings showed that ArtGAN-AEM is able to learn artistic representations from the Wikiart paintings that are usually non-figurative and abstract. For future work, we are looking forward to extend the work for other interesting applications, such as natural to artistic image translation based on a desired semantic-level mode, \eg \textit{style}.

\section*{Acknowledgment}
We gratefully acknowledge the support of NVIDIA Corporation with the donation of the Titan X GPU used for this research.

\ifCLASSOPTIONcaptionsoff
  \newpage
\fi

\bibliographystyle{IEEEtran}
\bibliography{ref}

\begin{IEEEbiography}[{\includegraphics[width=1in,height=1.25in,clip,keepaspectratio]{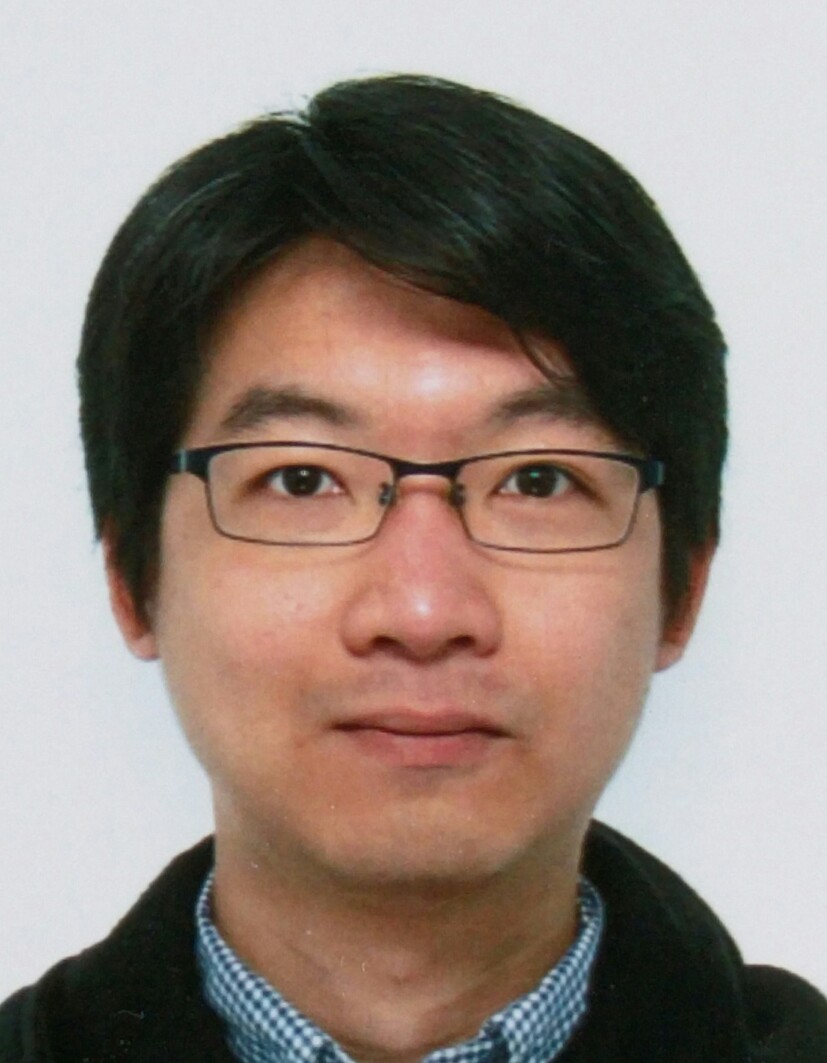}}]{Wei Ren Tan} received his Bachelor’s and Master’s degree in computer science from University of Malaya in Kuala Lumpur, Malaysia, in 2010 and 2013. He received his Doctor of Engineering degree from Shinshu University, Japan, in 2017. Currently, he is a Postdoctoral Research Fellow in the Computer and Communication Research Center of National Tsing Hua University, Taiwan. His research interests include computer vision, machine learning, and deep learning, focusing on image and video analysis.
\end{IEEEbiography}
\begin{IEEEbiography}[{\includegraphics[width=1in,height=1.25in,clip,keepaspectratio]{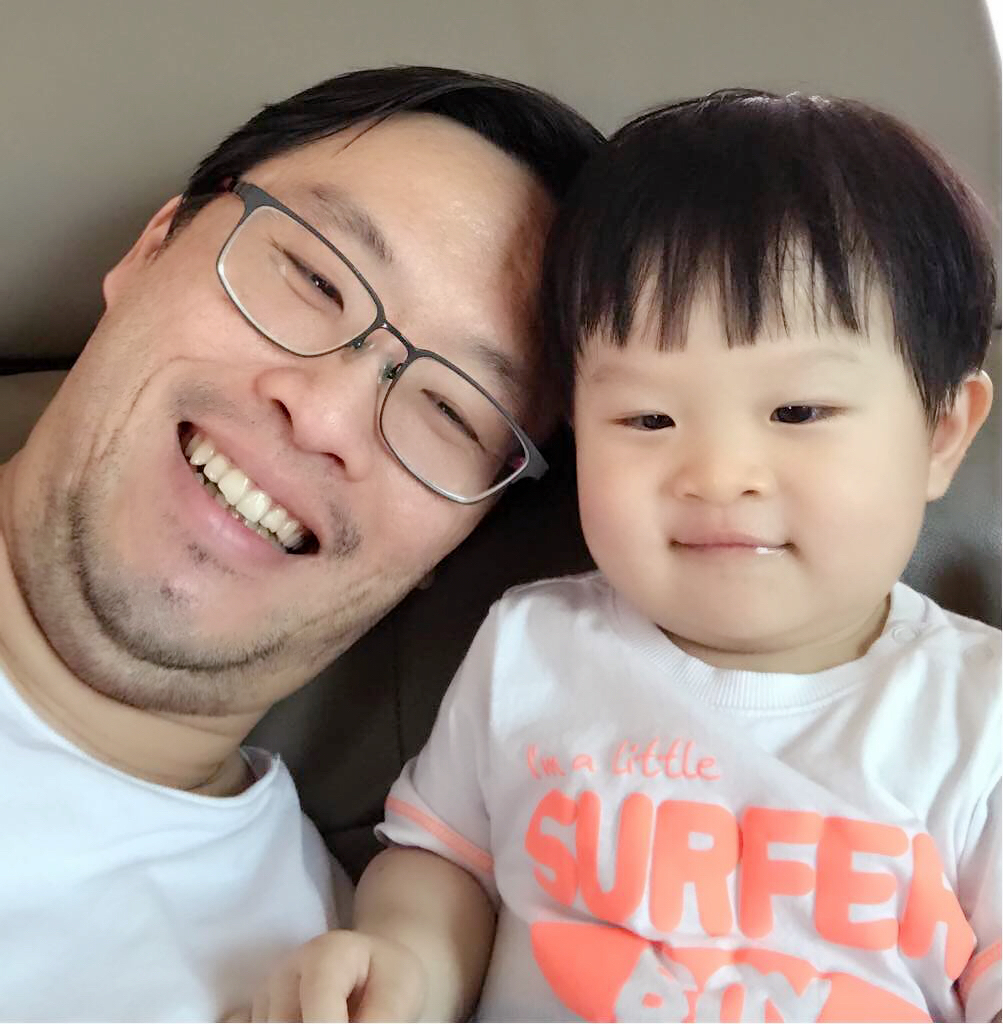}}]{Chee Seng Chan} (S'05-M'09-SM'14) received his Ph.D. degree from the University of Portsmouth, U.K., in 2008. He is currently an Associate Professor with the Faculty of Computer Science and Information Technology, University of Malaya, Malaysia. His research interests include computer vision and fuzzy set theory, particularly on image/video content analysis. He received several notable awards, such as the Young Scientist Network-Academy of Sciences Malaysia in 2015 and the Hitachi Research Fellowship in 2013. He is/was the Founding Chair of the IEEE Computational Intelligence Society Malaysia Chapter, the Organizing Chair of the Asian Conference on Pattern Recognition (ACPR2015), and the General Chair of the IEEE International Workshop on Multimedia Signal Processing (MMSP2019) and IEEE Visual Communications and Image Processing (VCIP2013).
\end{IEEEbiography}
\begin{IEEEbiography}[{\includegraphics[width=1in,height=1.25in,clip,keepaspectratio]{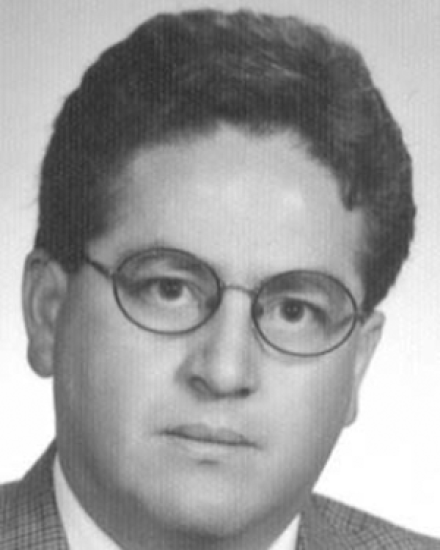}}]{Hernan Aguirre} received the engineer degree in computer systems from Escuela Politécnica Nacional, Quito, Ecuador, and the M.S. and Ph.D. degrees from Shinshu University, Nagano, Japan, in 1992, 2000, and 2003, respectively. He is an Associate Professor with Shinshu University. His research interests include evolutionary computation, multidisciplinary design optimization, and sustainability. He has authored over 130 international journal and conference research papers in related areas. Dr. Aguirre is a member of the IEICE and IPSJ.
\end{IEEEbiography}
\begin{IEEEbiography}[{\includegraphics[width=1in,height=1.25in,clip,keepaspectratio]{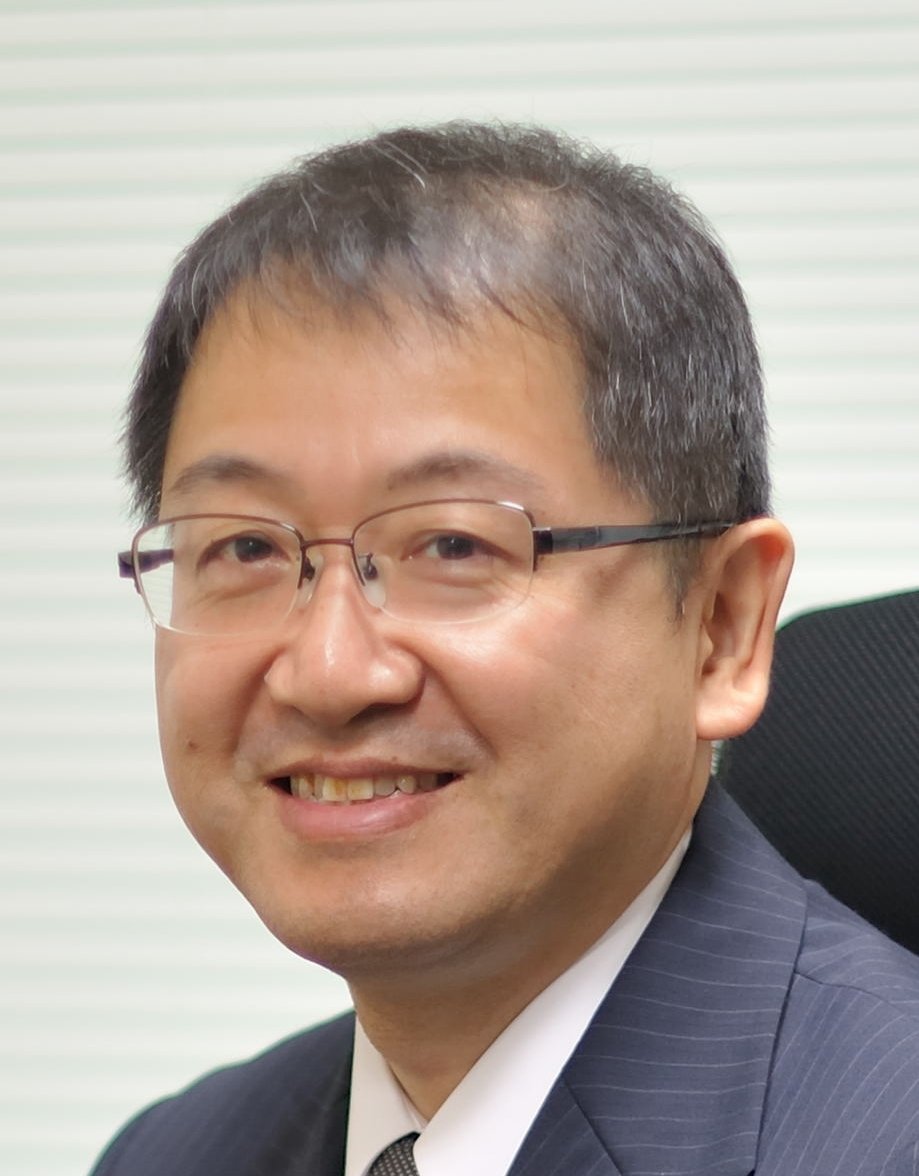}}]{Kiyoshi Tanaka} received his B.S and M.S. degrees in Electrical Engineering and Operations Research from National Defense Academy, Yokosuka, Japan, in 1984 and 1989, respectively. In 1992, he received Dr. Eng. degree from Keio University, Tokyo, Japan. Currently he is a full professor at Shinshu University, Nagano, Japan. He is also the director of Global Education Center and Vice-President of Shinshu University. His research interests include image and video processing, 3D point cloud processing, information hiding, human visual perception, evolutionary computation, multi-objective optimization, smart grid, and their applications. He is a fellow of IIEEJ and member of IEEE, IEICE, IPSJ and JSEC.
\end{IEEEbiography}
\vfill

\end{document}